\documentclass{article} % For LaTeX2e
\usepackage{iclr2026_conference,times}

\usepackage{amsmath,amsfonts,bm}

\def\eqref#1{equation~\ref{#1}}
\def\1{\bm{1}}

\DeclareMathAlphabet{\mathsfit}{\encodingdefault}{\sfdefault}{m}{sl}
\SetMathAlphabet{\mathsfit}{bold}{\encodingdefault}{\sfdefault}{bx}{n}

\usepackage{url}
\usepackage{amsmath}
\usepackage{amsfonts}
\usepackage{amssymb}
\usepackage{mathtools}
\usepackage{graphicx}
\usepackage{booktabs}
\usepackage{float}
\usepackage{array}
\usepackage{multirow}
\usepackage{enumitem}
\usepackage{hyperref}

\title{Task-Aware QUBO Allocation for Mixed-Precision Quantization}

\author{%
  Osama Orabi$^{1,2,4}$, Artur Zagitov$^{1}$, Hadi Salloum$^{1,2,3}$, Viktor A. Lobachev$^{4}$ \& Yaroslav Kholodov$^{1,4,5}$ \\
  $^{1}$Laboratory of Quantum Computing, Innopolis University, 420500 Innopolis, Russia\\
  $^{2}$Q Deep, Laboratory of Quantum Computing, 420502 Innopolis, Russia\\
  $^{3}$Research Center of the Artificial Intelligence Institute, Innopolis University, 420500 Innopolis, Russia\\
  $^{4}$Moscow Independent Research Institute of Artificial Intelligence (MIRAI), 117218 Moscow, Russia\\
  $^{5}$Sirius University of Science and Technology, 354340 Sochi, Russia
}

\iclrfinalcopy

\begin{document}

\maketitle

\begin{abstract}
Mixed-precision quantization requires discrete allocation of weight and
activation bit-widths, followed by recovery of the selected network.
We develop a task-aware quadratic unconstrained binary optimization (QUBO)
surrogate with separate weight and activation profiles, a bit-operation
(BOP) cost, and selected structural priors. QUBO provides a network-wide
allocation that can be refined through direct validation-based PROTES search.
On a compact NAFBlock-based denoiser, the refined route achieves 37.192 dB
after LSQ+ at 4.035\% routed-layer BOPs, versus 37.092 dB at 4.101\% for a
HAWQ-style baseline. The repeated-search primary experiment shows that LSQ+
largely closes the quality gap between QUBO allocation and expensive direct
refinement. An additional restoration architecture retains a larger recovered
gain, indicating that refinement's value depends on architecture and recovery.
We evaluate quality, achieved cost, routing stability and optimization expense
together. Deployment measurements characterize a fake-quantized floating-point
implementation; BOP reductions describe analytical allocation savings.
\end{abstract}

\section{Introduction}

Mixed-precision quantization assigns different numerical precisions to
network components to balance task quality and resource cost. Methods such
as HAQ and HAWQ use hardware feedback or sensitivity estimates to guide
allocation~\cite{wang2019haq,dong2019hawq,dong2020hawq}. Learned quantizers,
including LSQ and LSQ+~\cite{esser2019learned,bhalgat2020lsq+}, adapt scales
and offsets once precisions are fixed. The allocation and recovery problems
are therefore related but distinct: a route that performs well before
recovery need not remain the best route afterward.

A tractable allocation surrogate must also distinguish weights from
activations. Local weight reconstruction error provides a candidate-specific
signal, but intermediate activation perturbations affect subsequent
computation. In the studied restoration network, unrestricted activation menus produce
severely degraded routes under the tested QUBO search, motivating the
empirical activation profile and asymmetric candidate space.
Selected structural dependencies can further influence how these errors
combine. These observations motivate separate sensitivity models and
interaction terms rather than a layer-independent compression reward.

We combine a global discrete surrogate with a second stage that evaluates
candidate networks directly. The first stage uses QUBO to trade normalized
quantization damage against an analytical BOP cost. Hand-selected activation
coupling and module-order damage terms incorporate structural priors. The
second stage initializes PROTES, a tensor-train probabilistic optimizer,
from the best observed QUBO route~\cite{batsheva2023protes,oseledets2011tensor}.
A soft compute-growth penalty discourages refinement from purchasing quality
through a large increase in precision.

Our contribution is a task-aware QUBO allocation model and an empirical
assessment of when direct network search adds value. Separate activation
profiling and structural priors make the surrogate useful for joint routing;
QUBO-seeded PROTES tests the remaining gap to direct task evaluation. We
evaluate three distinct outcomes:
allocation quality before recovery, quality after recovery, and measured
execution behavior. Our evaluation includes a HAWQ-style sensitivity--knapsack baseline,
comparisons with stochastic search, validation-subset and solver robustness,
and additional restoration and classification architectures.

A central finding is that recovery changes the return on allocation search.
In the primary repeated-search experiment, QUBO provides a strong starting
route and LSQ+ leaves little additional gain for direct refinement. On NAFNetSmall, refinement retains a larger post-recovery benefit.
This identifies a practical trade-off between quality, recovery and search expense.

\section{Related Work}

\paragraph{Quantization and allocation.}
Post-training methods such as AdaRound and BRECQ optimize rounding or
reconstruction without full retraining~\cite{nagel2020up,li2021brecq}.
PACT, LSQ and LSQ+ learn quantizer parameters during
training~\cite{choi2018pact,esser2019learned,bhalgat2020lsq+}.
HAQ uses reinforcement learning for allocation, while HAWQ variants exploit
Hessian sensitivity~\cite{wang2019haq,dong2019hawq,dong2020hawq}.
Differentiable precision search, including EdMIPS, instead relaxes discrete
choices~\cite{cai2020edmips,wang2020differentiable}.
HAWQ-V3 further addresses integer-only execution and hardware-aware
allocation~\cite{yao2021hawqv3}. Our evaluated HAWQ-style
baseline combines Hessian weight sensitivity with the shared activation
profile and a multiple-choice allocator. It evaluates this allocation recipe
within our common quantization and recovery implementation. Uniform
allocation isolates the value of layer-dependent precision, while the
HAWQ-style allocator compares a Hessian-based allocation rule under the
same activation profile and recovery stack. This shared stack controls
implementation differences, but does not evaluate the complete HAWQ-V3
deployment system or establish superiority over differentiable search.

\paragraph{Discrete surrogates and direct search.}
QUBO represents binary allocation decisions and pairwise
interactions~\cite{kochenberger2014unconstrained,glover2018tutorial}; related
neural-network compression formulations have also been
studied~\cite{subinas2025optimization,wang2026quantum}.
Mu\~niz Subi\~nas et al. optimize per-parameter rounding at a fixed
precision, whereas Wang et al. study joint pruning and quantization using
adiabatic quantum computing. Our binary decisions instead select
layer-associated bit-widths, followed by classical direct refinement.
We solve the surrogate with classical simulated
annealing~\cite{kirkpatrick1983optimization}. PROTES optimizes a tensor-train sampling distribution
using black-box function evaluations~\cite{batsheva2023protes}.
Our distinction is its use after a sensitivity-based QUBO allocation, with
direct task evaluation and an explicit compute-growth penalty. The purpose
of this combination is to correct surrogate mismatch; its usefulness must
be assessed against both simpler allocation and unseeded search.

\section{Sensitivity-Aware Allocation and Direct Refinement}
\label{sec:method_main}

\begin{figure}[tbp]
\centering
\includegraphics[width=\linewidth]{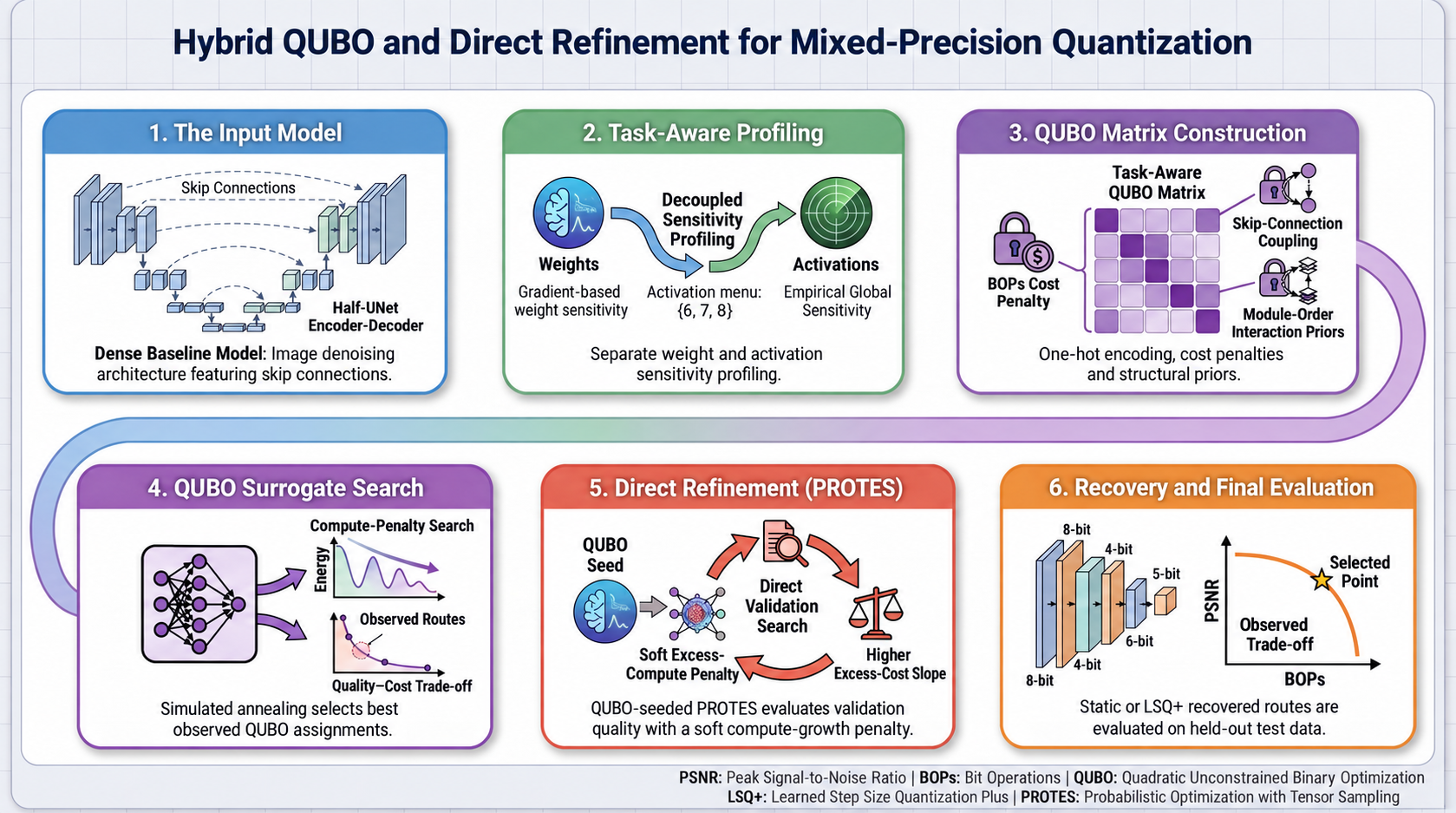}
\caption{Hybrid QUBO allocation and direct validation refinement. The input
network and trade-off sketches are schematic, not measured curves or a
complete architectural specification. Weight and activation profiling,
QUBO surrogate search, PROTES refinement and LSQ+ recovery are distinct
stages; both search stages are heuristic. The activation menu shown is the
final joint-quantization menu. Static or recovered routes are evaluated on
held-out test data.}
\label{fig:pipeline}
\end{figure}

Figure~\ref{fig:pipeline} summarizes the allocation and recovery stages.
We first construct a sensitivity-based discrete surrogate, then refine its
selected route using direct validation evaluation.

\paragraph{Candidate damage.}
For each routed convolution $n$, we select a weight precision
$b\in\mathcal C^W=\{4,5,6,7,8\}$ and an input-activation precision
$a\in\mathcal C^A=\{6,7,8\}$. Let $q^W_{n,b}$ and $q^A_{n,a}$ be binary
one-hot variables. The candidate damage coefficients are
\begin{equation}
d^W_{n,b}=F^W_n\epsilon^W_{n,b},\qquad
 d^A_{n,a}=F^A_n\epsilon^A_{n,a}.
\end{equation}
Here $\epsilon^W$ and $\epsilon^A$ are tensor reconstruction MSEs under
candidate quantization. $F^W$ is a square-root/max-normalized exponential
moving average of squared weight-gradient sums. It is Fisher-inspired, not
an exact Fisher trace. $F^A$ is a square-root/max-normalized score obtained
by temporarily quantizing one convolution output and evaluating its final
network effect. The final activation score differs from the local activation-gradient
proxy used in preliminary Phase 2 (Section~\ref{sec:development_main}).
Candidate activation MSE instead samples the convolution
input; these are associated but distinct tensor locations.

We preserve the profiling variants used by the experiments. Development
all-convolution studies measure probe reconstruction MSE against ground
truth. Additional-architecture studies measure disturbance from FP32 output
using MSE for restoration and KL divergence for classification. Development
weight candidate errors use a scale grid, while the reviewer reconstruction
and additional architectures use isolated 50-step scale calibration. Neither
is network-level LSQ+ recovery. Appendix~\ref{app:formulation} specifies the
normalization, quantizers and variants in detail.

\paragraph{Choice of activation profile.}
Preliminary studies used shared weight-gradient scores, then separate local
activation gradients (Phase 2). Neither the latter nor an activation scalar
resolved degraded routes. The final design uses downstream empirical
profiling and an asymmetric menu. Tables~\ref{tab:weight_progression}
and~\ref{tab:joint_progression} summarize these comparisons;
Appendix~\ref{sec:development_record} retains their chronology and protocols.

\paragraph{QUBO objective.}
Collecting the binary variables into $q$, the surrogate combines
\begin{equation}
E(q)=D(q)+\gamma C(q)+\omega R(q)+\alpha P(q),
\label{eq:main_qubo}
\end{equation}
where $D$ sums candidate damage after common maximum normalization,
$C$ is the weight--activation compute coupling, $R$ combines normalized
activation-damage products for consecutive convolutions in module traversal
order, and $P$ contains one-hot and selected activation-mismatch penalties.
The default damage scale is one and $\omega=0.15$ in the main formulation.
Equation~\ref{eq:main_qubo} groups the implemented matrix contributions;
Appendix~\ref{app:formulation} gives their exact coefficient construction.
The one-hot penalty is equivalent, up to a constant, to
$\sum_g(\sum_{u\in g}q_u-1)^2$. Selected mismatched activation choices
receive an additional exclusion penalty.

\paragraph{Matrix assembly.}
For an upper-triangular QUBO, initialize $Q_{uu}=\widehat d_u-\alpha$,
where $\widehat d_u=d_u/\max_v d_v$. Add $2\alpha$ to competing
one-hot pairs and selected skip-mismatch pairs, $\gamma C_{uv}$ to
weight--activation cost pairs, and $\omega\widehat d_u^A\widehat d_v^A$
to module-order activation pairs. Contributions to the same pair sum; all
other entries are zero. Appendix~\ref{app:formulation} specifies the
adaptive exclusion scale and coefficient definitions.

These are hand-selected priors: coupled convolution-input variables proxy
fusion paths, and module traversal need not follow data-flow edges. They
provide neither graph-derived interactions nor an integer fusion interface.

We anneal the surrogate and retain the best observed route. The extended
implementation repairs one-hot selections and performs up to four
coordinate-improvement passes; preliminary raw-solver studies remain a
separate variant. Gamma search seeks a requested compute level but discrete
routes can miss it. The normalized routed-layer cost is
\begin{equation}
B(\mathcal A)=
\frac{\sum_{n\in\mathcal R}\mathrm{MAC}_n b^W_n b^A_n}
{32^2\sum_{n\in\mathcal R}\mathrm{MAC}_n},
\label{eq:main_bops}
\end{equation}
with routed convolutions $\mathcal R$. Protected convolutions and other
floating-point operations are excluded. BOP percentages below refer to this
proxy, not whole-network latency or energy.

\paragraph{QUBO-seeded PROTES.}
PROTES samples candidate routes from a tensor-train probability model
initialized around the QUBO route. Restoration candidates are evaluated on
32 validation patches. With seed cost $B_0$, refinement maximizes
\begin{equation}
S(\mathcal A)=M(\mathcal A)-\gamma B(\mathcal A)
 -(h-\gamma)\max\{B(\mathcal A)-B_0,0\},
\quad h=\max\{10\gamma,100\},
\label{eq:main_refinement}
\end{equation}
where $M$ is validation PSNR or classification accuracy in percentage points.
The continuous score charges excess compute at the higher rate;
its soft penalty permits routes above the seed cost. The final
route is evaluated on the held-out test set, separately before and after
LSQ+. Full-sweep and reviewer experiments use different budgets and cache
representations; the appendix reports them separately.

The excess-cost term raises the marginal compute charge from $\gamma$ to
$h$ above the seed cost, while retaining the original charge for cheaper
routes. In preliminary runs, a small linear penalty allowed substantial
compute expansion; the added tax encouraged quality gains at similar or
lower cost (Table~\ref{tab:penalty_main}). This motivates the continuous
score rather than a hard feasibility constraint.

\section{Experimental Protocol}
\label{sec:protocol_main}

\paragraph{Evidence scope.}
The primary sweep reports individual recoveries; repeated-search studies
measure search variation conditional on a fixed recovery seed. Static
all-convolution ablations test formulation and solver sensitivity.
Additional models use separate checkpoints and one recovery replicate.
Table~\ref{tab:protocol_variants} maps these variants to their splits
and routing settings; results are interpreted within each protocol.

The primary model is a custom width-32 HalfUNet denoiser using
NAFBlocks~\cite{chen2022simple} and additive multi-scale fusion.
SIDD Small provides 160 RGB noisy/clean image
pairs~\cite{abdelhamed2018high}. Four quadrants per image are resized to
$256\times256$, producing 640 patches, then split into 512/64/64
train/validation/test patches with seed 42. Training exposes three accesses
per patch with flip augmentation. The primary batch size is 16. Because
splitting follows patch extraction, source images can cross partitions.
All reported primary SIDD scores use this resized-patch evaluation.

The main comparison retains full precision at the input/output boundaries
by protecting the first and last convolutions. It routes 38 internal
convolutions and uses 304 binary variables. The all-convolution statistical
and objective-development studies also route those boundary layers, giving
40 convolutions and 320 variables. Table~\ref{tab:protocol_variants}
identifies the setting used by each study. Additional restoration models are trained on
128/16/16 source-image directories; their patches cannot cross partitions,
but physical-scene IDs overlap. MobileNetV2~\cite{sandler2018mobilenetv2}
is evaluated on CIFAR-10. Full training and split settings appear in
Appendix~\ref{app:experiments}.

Static weights use a minimum-MSE symmetric scale search: the preliminary
evaluator uses an 80-point grid and the additional experiments
use 64 points. Table~\ref{tab:protocol_variants} preserves these
evaluation settings. Activations use dynamic symmetric
quantization. Convolutions, biases,
accumulation, normalization and residual operations remain floating point.
Extended LSQ+ recovery uses four quantizer-only epochs followed by fourteen
epochs updating quantizers and convolution parameters, capped at 100 batches
per epoch. Restoration uses L1 loss and classification cross-entropy.

Restoration outputs are clipped to $[0,1]$. The shared evaluator averages
PSNR computed from each batch's MSE and reports SSIM~\cite{wang2004image}.
The paired statistical analysis instead computes per-patch PSNR.
Dynamic activation scales include the batch dimension, so batch size and
aggregation are part of the evaluation protocol. Routing and calibration use validation data.

\section{Results}
\label{sec:results_main}

We report quality before and after recovery, the incremental benefit and
cost of direct search, structural ablations, and measured execution behavior.
The primary compute comparisons use achieved routed-layer BOPs; preliminary
approximate costs and missing costs are identified explicitly.

\subsection{Allocation Quality Before and After Recovery}

\begin{figure}[tbp]
\centering
\includegraphics[width=\linewidth]{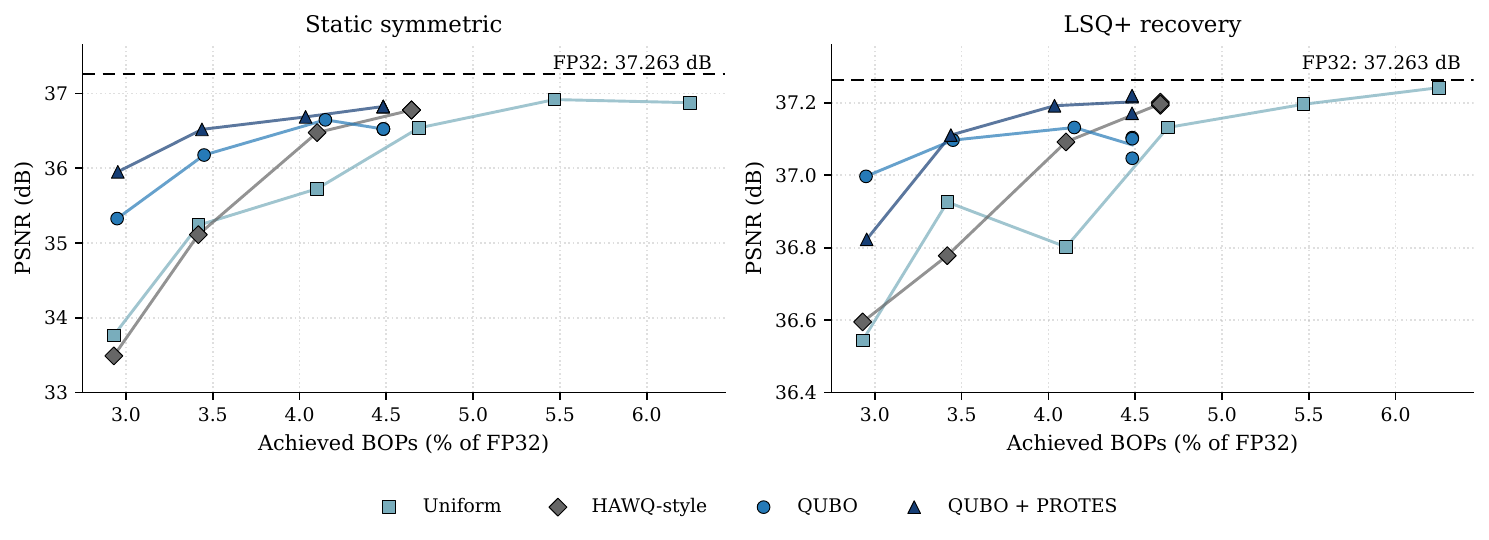}
\caption{HalfUNet weight--activation quality--compute trade-off from all six
requested targets in Table~\ref{tab:main_comparison}. Markers show the original
recorded evaluations at achieved routed-layer BOPs. The three loose targets
reuse one route per non-uniform method; separate recovered scores are shown
at their shared cost. Lines connect records to visualize the quality--compute trajectory, using
the mean at repeated costs. Markers retain the individual evaluations and
achieved costs. The two panels show how recovery changes route ordering.}
\label{fig:wa_frontier}
\end{figure}

Table~\ref{tab:main_comparison} reports all six requested targets, including
repeated routes. Figure~\ref{fig:wa_frontier} visualizes the same records
before and after recovery. Static and recovered records are paired by
assignment ID; each displayed quality cell is a recorded evaluation, and
reused routes are identified in the full table.

\begin{table}[tbp]
\centering
\small
\setlength{\tabcolsep}{4pt}
\caption{Primary patch-random HalfUNet comparison. Costs are routed-layer
BOP percentages relative to FP32; quality cells give PSNR (dB) / SSIM.
The six blocks, top to bottom, retain the requested targets 6.250,
5.469, 4.688, 4.102, 3.418 and 2.930\%. Only achieved BOPs are tabulated.
Bold marks each metric maximum within a block, excluding FP32; costs
differ across methods. Reused routes retain separate recovery evaluations.}
\label{tab:main_comparison}
\begin{tabular}{lrrr}
\toprule
Method & Achieved BOPs (\%) & Static PSNR / SSIM & LSQ+ PSNR / SSIM \\
\midrule
FP32 reference & 100.000 & 37.263 / 0.9328 & -- \\
\midrule
Uniform & 6.250 & \textbf{36.877} / \textbf{0.9245} & \textbf{37.242} / \textbf{0.9316} \\
HAWQ-style & 4.645 & 36.779 / 0.9224 & 37.202 / 0.9298 \\
QUBO & 4.483 & 36.525 / 0.9178 & 37.104 / 0.9295 \\
QUBO + PROTES & 4.482 & 36.826 / 0.9237 & 37.171 / 0.9308 \\
\midrule
Uniform & 5.469 & \textbf{36.919} / \textbf{0.9254} & 37.196 / 0.9308 \\
HAWQ-style & 4.645 & 36.779 / 0.9224 & 37.193 / 0.9294 \\
QUBO & 4.483 & 36.525 / 0.9178 & 37.047 / 0.9290 \\
QUBO + PROTES & 4.482 & 36.826 / 0.9237 & \textbf{37.217} / \textbf{0.9311} \\
\midrule
Uniform & 4.688 & 36.540 / 0.9199 & 37.132 / 0.9305 \\
HAWQ-style & 4.645 & 36.779 / 0.9224 & 37.197 / 0.9299 \\
QUBO & 4.483 & 36.525 / 0.9178 & 37.101 / 0.9295 \\
QUBO + PROTES & 4.482 & \textbf{36.826} / \textbf{0.9237} & \textbf{37.220} / \textbf{0.9309} \\
\midrule
Uniform & 4.102 & 35.727 / 0.9026 & 36.802 / 0.9242 \\
HAWQ-style & 4.101 & 36.477 / 0.9152 & 37.092 / 0.9278 \\
QUBO & 4.149 & 36.648 / 0.9215 & 37.132 / 0.9298 \\
QUBO + PROTES & 4.035 & \textbf{36.687} / \textbf{0.9221} & \textbf{37.192} / \textbf{0.9308} \\
\midrule
Uniform & 3.418 & 35.238 / 0.8936 & 36.926 / 0.9268 \\
HAWQ-style & 3.417 & 35.112 / 0.8810 & 36.778 / 0.9215 \\
QUBO & 3.450 & 36.177 / 0.9102 & 37.097 / 0.9292 \\
QUBO + PROTES & 3.438 & \textbf{36.520} / \textbf{0.9193} & \textbf{37.111} / \textbf{0.9294} \\
\midrule
Uniform & 2.930 & 33.765 / 0.8552 & 36.544 / 0.9178 \\
HAWQ-style & 2.930 & 33.490 / 0.8464 & 36.595 / 0.9175 \\
QUBO & 2.949 & 35.327 / 0.9058 & \textbf{36.997} / \textbf{0.9279} \\
QUBO + PROTES & 2.953 & \textbf{35.952} / \textbf{0.9113} & 36.823 / 0.9239 \\
\bottomrule
\end{tabular}
\end{table}

At the requested 4.102\% target, refinement achieves 37.192 dB at 4.035\%
BOPs, versus 37.132 dB at 4.149\% for QUBO and 37.092 dB at 4.101\%
for the HAWQ-style allocator. At the requested 3.418\% target, QUBO reaches
37.097 dB at 3.450\%, compared with 36.778 dB at 3.417\% for HAWQ-style.
These are nearby-cost observations, not exact matched-cost significance tests.

Recovery changes the ordering of routes. At the tightest target, refinement
improves static PSNR from 35.327 to 35.952 dB but reduces recovered PSNR
from 36.997 to 36.823 dB. At the three looser targets, QUBO and refined
routes are reused, with separate recovery evaluations. Uniform quantization
is competitive at larger costs. The original uniform recovery score at
4.102\% is 36.802 dB; a later run on the same assignment records 37.010 dB.
The difference illustrates why isolated post-recovery gaps should be scoped
as observations rather than reliable method-level effects.

\subsection{Additional Architectures and Tasks}

Table~\ref{tab:generalization_main} reports recovered quality on source-image-disjoint SIDD restoration and CIFAR-10 classification. These are separate
training settings from the primary checkpoint, each with one recorded
recovery seed. Full static
results and training details appear in Appendix Table~\ref{tab:generalization}.

\begin{table}[t]
\centering
\small
\caption{Additional architectures at a requested 4.102\% BOP target.
Achieved cost is routed-layer BOP percentage. Recovery quality is PSNR / SSIM
for restoration and accuracy (\%) for classification; one recorded run per cell.
FP32 rows show dense-checkpoint quality without recovery.
SIDD partitions share physical scenes. Training and evaluation settings
appear in Appendix~\ref{sec:generalization}.}
\label{tab:generalization_main}
\begin{tabular}{lcc}
\toprule
Method & Achieved BOPs (\%) & Quality \\
\midrule
\multicolumn{3}{l}{\textit{NAFNetSmall / SIDD (PSNR / SSIM)}} \\
FP32 reference & 100.000 & 34.743 / 0.8710 \\
Uniform & 4.102 & 32.865 / 0.8134 \\
HAWQ-style & 4.111 & 33.571 / 0.8461 \\
QUBO & 4.113 & 33.703 / 0.8429 \\
QUBO + PROTES & 4.107 & 34.079 / 0.8514 \\
\midrule
\multicolumn{3}{l}{\textit{MobileNetV2 / CIFAR-10 (accuracy, \%)}} \\
FP32 reference & 100.000 & 91.720 \\
Uniform & 4.102 & 91.770 \\
HAWQ-style & 3.580 & 91.760 \\
QUBO & 3.493 & 91.680 \\
QUBO + PROTES & 3.475 & 91.860 \\
\bottomrule
\end{tabular}
\end{table}

On NAFNetSmall, refinement changes recovered PSNR from 33.703 to
34.079 dB at approximately 4.11\% BOPs. MobileNetV2 changes
from 91.68\% to 91.86\% after recovery, while its static accuracy decreases.
Uniform allocation also attains the highest recorded quality at some larger
budgets. The NAFNetSmall gain exceeds that of the primary repeated-search
experiment. The small single-run classification differences provide a
limited cross-task check, without establishing a general accuracy advantage.

\subsection{Does QUBO Initialization Help Direct Search?}

We compare random search, randomly initialized PROTES, network-score
simulated annealing and QUBO-seeded PROTES on one 32-patch validation subset,
with seeds 1701--1703 and a nominal budget of 2,000 proposals. We also
recover and evaluate the fixed QUBO seed; its recovery evaluation is separate
from the six-target sweep. Table~\ref{tab:search_main} gives
quality and achieved mean cost; the appendix reports realized counts.

\begin{table}[t]
\centering
\small
\caption{Search comparison at one requested compute target. Quality values
are test-set means and, for the stochastic methods, standard deviations over
three search seeds. QUBO alone is a single fixed assignment. The BOP column
reports achieved mean cost. Unique evaluations are rounded means over
search seeds; LSQ+ variation is conditional on one fixed recovery seed.}
\label{tab:search_main}
\setlength{\tabcolsep}{4pt}
\begin{tabular}{lrrrr}
\toprule
Method & BOPs (\%) & Static PSNR (dB) & LSQ+ PSNR (dB) & Unique evals. \\
\midrule
QUBO alone & 4.149 & 36.648 & 37.153 & 1 \\
QUBO-seeded PROTES & 4.129 & $36.694\pm0.081$ & $37.160\pm0.015$ & 1,953 \\
Random PROTES & 4.079 & $36.550\pm0.026$ & $37.040\pm0.082$ & 2,000 \\
Random search & 4.094 & $36.074\pm0.158$ & $37.020\pm0.037$ & 2,000 \\
Simulated annealing & 4.143 & $36.617\pm0.109$ & $37.137\pm0.019$ & 1,700 \\
\bottomrule
\end{tabular}
\end{table}

Seeded PROTES has the highest mean static score among the tested stochastic
methods, but its improvement over QUBO alone is 0.046 dB before LSQ+ and
0.007 dB afterward. Achieved costs also differ slightly. The mean realized
unique evaluations are approximately 1,953 for seeded PROTES, 2,000 for
random PROTES and random search, and 1,700 for network-score SA. These are
comparable nominal budgets, not identical evaluation counts.

\subsection{Effect of Structural Priors}

Table~\ref{tab:topological_ablation} reports preliminary structural comparisons
at $\gamma=50$ and $200$. At $\gamma=50$, coupling raises PSNR from 35.105
to 35.386 dB while increasing cost from 3.603\% to 4.157\%. At $\gamma=200$,
adding module-order interactions improves
static PSNR by 0.446 dB relative to activation coupling alone at the same
reported cost. The uncoupled comparison also changes cost. These single-run
preliminary observations are separate from the protected-layer recovered
comparison; Appendix~\ref{sec:topological_ablation} retains the wider sweep.

\begin{table}[tbp]
\centering
\small
\setlength{\tabcolsep}{4pt}
\caption{Development structural-prior comparisons before LSQ+. Average bits
are preliminary layer averages. Only the two coupled $\gamma=200$ rows have
identical reported BOPs. Single-run records do not establish recovered
multi-seed effects or an optimal interaction coefficient.}
\label{tab:topological_ablation}
\begin{tabular}{rlrrrr}
\toprule
$\gamma$ & Formulation & Avg. W / A & BOPs (\%) & PSNR & SSIM \\
\midrule
50 & No structural priors & 5.58 / 7.17 & 3.603 & 35.105 & 0.887 \\
50 & Activation coupling & 6.05 / 7.00 & 4.157 & 35.386 & 0.894 \\
\midrule
200 & No structural priors & 5.90 / 7.12 & 3.568 & 34.368 & 0.869 \\
200 & Activation coupling & 5.53 / 7.05 & 3.374 & 34.455 & 0.870 \\
200 & Coupling + $\omega=0.15$ & 5.55 / 7.00 & 3.374 & 34.901 & 0.891 \\
\bottomrule
\end{tabular}
\end{table}

\subsection{Formulation Comparisons}
\label{sec:development_main}

The preliminary comparisons in Tables~\ref{tab:weight_progression}
and~\ref{tab:joint_progression} motivate the final damage model.
Task-aware weighting improves the one-hot weight-only route from 36.803 to
36.925 dB, versus 36.627 dB for uniform six-bit weights. For joint allocation,
separate local activation-gradient scores yield 25.481 dB; the later empirical
profile with the asymmetric menu reaches 35.105 dB at 3.603\% BOPs.
The latter route uses 5.58-bit weights and 7.17-bit activations. Profiling and
the menu changed together, so this comparison evaluates successive designs;
the isolated menu study appears in Appendix~\ref{sec:activation_cliff}.
Appendix~\ref{sec:development_record} gives the full phase chronology.

\begin{table}[!htb]
\centering
\small
\caption{Weight-only objective-development comparisons before LSQ+. Activations
are not jointly allocated; average precision and PSNR use the preliminary
evaluation protocol summarized in Appendix~\ref{sec:development_record}.}
\label{tab:weight_progression}
\begin{tabular}{lrr}
\toprule
Weight allocation & Avg. W bits & PSNR (dB) \\
\midrule
FP32 reference & 32.00 & 37.263 \\
Uniform eight-bit & 8.00 & 37.207 \\
Original bit-reduction QUBO & 7.94 & 35.778 \\
Uniform six-bit & 6.00 & 36.627 \\
One-hot reconstruction QUBO & 6.00 & 36.803 \\
One-hot task-aware QUBO & 6.03 & 36.925 \\
\bottomrule
\end{tabular}
\end{table}

\begin{table}[!htb]
\centering
\small
\setlength{\tabcolsep}{4pt}
\caption{Joint W+A formulation-development comparisons before LSQ+. Average bits
are layer averages; approximate BOPs are report estimates. Profiling and menus
vary across these preliminary settings (Appendix~\ref{sec:development_record});
the fixed-A8 route has no recorded achieved BOPs.}
\label{tab:joint_progression}
\begin{tabular}{lrrrrr}
\toprule
Strategy & Avg. W & Avg. A & BOPs (\%) & PSNR & SSIM \\
\midrule
Uniform W4/A8 & 4.00 & 8.00 & 3.125 & 33.405 & 0.886 \\
Uniform W6/A8 & 6.00 & 8.00 & 4.688 & 36.297 & 0.917 \\
Shared weight-gradient score & 6.15 & 6.15 & $\sim$3.75 & 33.750 & 0.866 \\
Separate local W/A gradients & 6.10 & 6.00 & $\sim$3.60 & 25.481 & 0.447 \\
Local gradients + A scalar & 5.62 & 6.50 & $\sim$3.55 & 24.896 & 0.431 \\
QUBO W + fixed A8 & 5.925 & 8.00 & -- & 36.448 & 0.917 \\
Empirical + asymmetric menu & 5.58 & 7.17 & 3.603 & 35.105 & 0.887 \\
\bottomrule
\end{tabular}
\end{table}

Table~\ref{tab:penalty_main} shows the effect of charging for compute growth
in the preliminary refinement study. At $\gamma=1.58$, linear-score refinement
raises BOPs from 3.990\% to 4.989\%, while expansion-penalized refinement
reaches 36.318 dB at 3.985\%, a 0.821 dB gain over its seed. The observations
motivate the excess-cost term in Eq.~\ref{eq:main_refinement}; the saved report's
piecewise tax description does not establish exact equivalence to the current
continuous implementation.

\begin{table}[!htb]
\centering
\small
\setlength{\tabcolsep}{5pt}
\caption{Refinement objective-development comparison before LSQ+, with a nominal
budget of 2,000 per run. Cells give achieved BOPs (\%) / PSNR (dB), starting
from the displayed seed. These preliminary records use the development
protocol, separately from Table~\ref{tab:search_main}.}
\label{tab:penalty_main}
\begin{tabular}{rccc}
\toprule
$\gamma$ & QUBO seed & Linear-score refinement & Expansion-penalized refinement \\
\midrule
85.77 & 3.323 / 35.144 & 3.113 / 35.573 & 3.165 / 35.704 \\
11.66 & 4.119 / 35.969 & 4.790 / 36.586 & 4.042 / 36.300 \\
1.58 & 3.990 / 35.497 & 4.989 / 36.460 & 3.985 / 36.318 \\
\bottomrule
\end{tabular}
\end{table}

\subsection{Robustness of Profiling and Routing}

Supporting studies identify several limits of stability. Five overlapping
32-patch validation subsets produce static scores of approximately
$36.712\pm0.037$ dB, but only 37--58\% pairwise agreement in both layer
precisions. Probe-bit ranking correlations are high (Spearman 0.936--0.984),
but this does not establish stable coefficient magnitudes or routes.
The all-convolution solver-seed study has PSNR standard deviations above
1 dB at two tested gamma values despite relatively stable surrogate energy.
Finally, a paired static QUBO--uniform comparison records a 0.3769 dB
mean improvement at modestly higher BOPs. Its nominal patch-level confidence
interval is [0.1748, 0.5790] dB; shared images/scenes limit independence.
Its scope is the static, all-convolution comparison, rather than recovered
quality under the primary protected-layer protocol.

Alternative Hessian-QUBO, entropy-based activation unlocking and sub-layer
routing designs did not consistently improve the evaluated configurations
(Appendix~\ref{sec:negative_results}). These observations motivate the
simpler final design without establishing that those alternatives are
universally inferior.

\subsection{Optimization Cost and Reference Implementation}

\begin{table}[tbp]
\centering
\small
\setlength{\tabcolsep}{5pt}
\caption{Allocation and recovery costs at the primary requested 4.102\% target.
End-to-end costs exclude dense training; recovery includes materialization.
These timings belong to the original six-target sweep, with a different
refinement budget from the repeated-search study in Table~\ref{tab:search_main}.}
\label{tab:cost_deployment_main}
\begin{tabular}{lrr}
\toprule
Method & End-to-end (s) & Recovery (s) \\
\midrule
Uniform & 185.3 & 164.8 \\
HAWQ-style & 205.6 & 178.0 \\
QUBO & 198.1 & 165.3 \\
QUBO + PROTES & 734.5 & 172.6 \\
\bottomrule
\end{tabular}
\end{table}

Table~\ref{tab:cost_deployment_main} quantifies the additional expense of
direct refinement in the six-target sweep. At this recorded operating point,
the end-to-end cost rises from 198.1 s for QUBO to 734.5 s for QUBO + PROTES.
The repeated-search comparison uses a separate 2,000-proposal budget;
its mean seeded-PROTES search runtime is 1,249.8 s, excluding recovery.
The quality and timing records therefore retain their respective protocols.

The reference implementation applies fake quantization and executes
floating-point convolutions. On an NVIDIA H100, batch-one inference with
$3\times256\times256$ inputs takes $3.002\pm0.094$ ms for FP32 and
$6.735\pm0.249$ ms for QUBO; quantized routes are near 6.7 ms after 25 warmups
and 200 timed repetitions. Serialized model size and peak GPU allocation also
remain unchanged or increase. These measurements document the overhead of
the reference implementation. Realizing the analytical BOP and packed-storage
reductions requires native low-bit kernels and packed tensors. Full CPU/GPU
measurements and environment details appear in Appendix~\ref{sec:hardware}.

\section{Discussion and Limitations}

Recovery changes the return on allocation search. QUBO provides a strong
starting route in the primary repeated-search experiment, where LSQ+ leaves
little additional gain for direct refinement. NAFNetSmall retains a larger recovered gain. In the studied primary setting,
QUBO alone is a practical choice when LSQ+ recovery is available and search
budget is limited; refinement merits additional budget when residual quality
gaps remain, as in NAFNetSmall. A direct study
of surrogate scores and recovered route rankings would further characterize
recovery-aware allocation.

The primary split shares source images; the additional SIDD split separates
images but shares physical scenes. Most final comparisons use one recovery
run. The HAWQ-style allocator is evaluated within our shared implementation;
differentiable mixed-precision baselines remain to be tested. The all-convolution
studies use a separate protocol, and exploratory objective sweeps report
test quality without independently validating hyperparameter selection.
Complete penalty/normalization sweeps and unrestricted-activation PROTES
are additional open comparisons.

The structural interactions are selected proxies. Routed-layer BOPs exclude
protected layers and nonconvolutional operations; ideal packed storage is
separate from serialized storage. Native low-bit kernels, packed tensors and
integer fusion interfaces are needed to realize deployment gains.
HAWQ-V3 demonstrates hardware speedups through integer-only TVM deployment
on T4 GPUs~\cite{yao2021hawqv3}; such results depend on supported kernels
and hardware and do not calibrate our BOP proxy to latency.

\section{Conclusion}

Task-aware QUBO produces effective joint weight--activation allocations
using empirical activation profiling and selected structural priors. Direct
PROTES refinement can improve static quality, while its post-recovery return
depends on the architecture. LSQ+ largely closes the refinement gap in the
primary repeated-search experiment, making the QUBO route a strong lower-cost
starting point. NAFNetSmall retains a larger recovered gain.
Together, these results make recovery and optimization expense part of the
allocation decision, alongside quality and achieved BOPs.

\bibliography{iclr2026_conference}
\bibliographystyle{iclr2026_conference}

\clearpage
\appendix

\section{Detailed QUBO Formulation}\label{app:formulation}

The proposed framework follows a sequential progression. First, layer
sensitivity and candidate quantization errors are measured. These quantities
are incorporated into a QUBO objective containing one-hot routing penalties,
weight--activation computational-cost coupling, skip-connection
compatibility terms, and adjacent-activation interaction terms. The best
observed QUBO configuration is subsequently used to initialize PROTES.

\subsection{Decision Variables and One-Hot Encoding}

Let $\mathcal{C}^{W}$ denote the candidate weight bit-widths and
$\mathcal{C}^{A}$ denote the candidate activation bit-widths. For each
selected layer $n$, binary variables are defined as

\begin{equation}
q_{n,b}^{W}\in\{0,1\},
\qquad
\forall b\in\mathcal{C}^{W},
\end{equation}

\begin{equation}
q_{n,a}^{A}\in\{0,1\},
\qquad
\forall a\in\mathcal{C}^{A}.
\end{equation}

The selected precisions are represented by

\begin{equation}
b_n^{W}
=
\sum_{b\in\mathcal{C}^{W}}
bq_{n,b}^{W},
\qquad
b_n^{A}
=
\sum_{a\in\mathcal{C}^{A}}
aq_{n,a}^{A}.
\end{equation}

The implemented QUBO assigns a large exclusion penalty to competing
precision states within the same routing group. For every conflicting
one-hot pair $(i,j)$,

\begin{equation}
Q_{ij}\mathrel{+}=2\alpha.
\end{equation}

The diagonal coefficient for each precision state contains a corresponding
negative exclusion contribution.

\subsection{Task-Aware Weight Quantization Damage}

For weight tensor $W_n$ and candidate weight precision $b$, the symmetric
quantization error is obtained using a grid search over the quantizer scale:

\begin{equation}
\epsilon_{n,b}^{W}
=
\min_{s\in\mathcal{S}_{n,b}}
\frac{1}{|W_n|}
\left\|
Q_b(W_n;s)-W_n
\right\|_2^2.
\end{equation}

For $b>1$, the reference scale is

\begin{equation}
s_0
=
\frac{
\max |W_n|
}{
2^{b-1}-1
},
\end{equation}

and candidate scales are generated by multiplying $s_0$ by factors uniformly
spaced between $0.05$ and $1.20$.

The preliminary grid-MSE formulation estimates candidate errors using a
64-point scale grid. The reviewer QUBO reconstruction, HAWQ-style candidate
construction and new architecture experiments instead use an LSQ-calibrated
weight-error variant: a temporary scale is optimized for 50 Adam steps at
learning rate $10^{-2}$, keeping the weight tensor fixed, and its final MSE
replaces the grid-search error. This local calibration is distinct from
network-level LSQ+ recovery. The preliminary final evaluator specifies an
80-point minimum-MSE scale grid. The expanded W+A static evaluations
and additional architectures use 64 points. These evaluator variants must be distinguished from
the 64-point candidate-error profiling grid. We distinguish these variants when
interpreting the preliminary ablations and expanded comparisons.

To account for differences in layer sensitivity, we accumulate the squared
weight gradients over calibration batches. Let

\begin{equation}
S_n^{W,(t)}
=
\sum_{i\in W_n}
\left(
\frac{\partial\mathcal{L}^{(t)}}{\partial w_i}
\right)^2
\end{equation}

denote the gradient-energy score for layer $n$ at calibration batch $t$.

The implementation aggregates this quantity using an exponential moving
average with $\eta=0.9$:

\begin{equation}
\widetilde{S}_n^{W,(t)}
=
\eta
\widetilde{S}_n^{W,(t-1)}
+
(1-\eta)
S_n^{W,(t)}.
\end{equation}

The EMA is initialized with the first batch score. After processing the
calibration batches, the square-root transformation and
global normalization are applied:

\begin{equation}
\mathcal{F}_n^W
=
\frac{
\sqrt{\widetilde{S}_n^W}
}{
\max_j
\sqrt{\widetilde{S}_j^W}
}
+
10^{-6}.
\end{equation}

We refer to $\mathcal{F}_n^W$ as the
\emph{normalized gradient-based layer sensitivity score}. This quantity is
Fisher-inspired but is not identified as the literal Fisher Information
trace.

The sensitivity-weighted weight damage for candidate bit-width $b$ is

\begin{equation}
L_{n,b}^{W}
=
\mathcal{F}_n^W
\epsilon_{n,b}^{W}.
\end{equation}

The aggregate weight contribution is

\begin{equation}
L_W
=
\sum_n
\sum_{b\in\mathcal{C}^{W}}
L_{n,b}^{W}q_{n,b}^{W}.
\end{equation}

\subsection{Empirical Global Activation Sensitivity}

The global activation profiling procedure perturbs the output of one
collected convolutional layer at a time using a temporary symmetric 4-bit
quantizer.

For activation tensor $z_n$, let

\begin{equation}
q_{\max}
=
2^{b_{\mathrm{test}}-1}-1,
\qquad
b_{\mathrm{test}}=4,
\end{equation}

and

\begin{equation}
s_n
=
\frac{\max |z_n|}{q_{\max}}.
\end{equation}

The perturbed activation is

\begin{equation}
\widetilde{z}_n
=
\operatorname{clip}
\left(
\operatorname{round}
\left(
\frac{z_n}{s_n}
\right),
-q_{\max},
q_{\max}
\right)s_n.
\end{equation}

For every layer, the modified network is evaluated on a validation subset.
The global activation damage score is the average final reconstruction MSE:

\begin{equation}
G_n^A
=
\frac{1}{|\mathcal{D}_p|}
\sum_{(x,y)\in\mathcal{D}_p}
\operatorname{MSE}
\left(
f_{n,4}(x),y
\right),
\end{equation}

where $f_{n,4}$ denotes the model with the temporary 4-bit output
quantization inserted at layer $n$.

The implementation applies a square-root transformation and normalizes the
scores:

\begin{equation}
\mathcal{F}_n^A
=
\frac{
\sqrt{G_n^A}
}{
\max_j\sqrt{G_j^A}
}
+
10^{-6}.
\end{equation}

This ground-truth-loss definition is used in the preliminary all-convolution
studies. In the additional architecture experiments, the global probe instead
measures disturbance relative to the unperturbed FP32 output: mean squared
output distance for restoration and KL divergence from the FP32 class
probabilities for classification. The same square-root/max normalization is
applied. These are distinct sensitivity variants, not interchangeable loss
measurements. The output probe and input-activation MSE are associated with
the same convolution, but perturb different tensor locations.

\subsection{Candidate-Specific Activation Quantization Error}

Candidate-specific activation error is measured separately from the global
activation sensitivity. Profiling captures the input activation to each
collected convolutional layer. For each candidate bit-width $a$, the
activation is quantized using the same symmetric scale-search procedure
described above, and its minimum MSE is recorded.

Let $z_n(x)$ denote the sampled input activation to layer $n$. The
candidate-specific activation error is

\begin{equation}
\epsilon_{n,a}^{A}
=
\frac{1}{|\mathcal{D}_a|}
\sum_{x\in\mathcal{D}_a}
\min_{s\in\mathcal{S}_{n,a}}
\operatorname{MSE}
\left(
Q_a(z_n(x);s),
z_n(x)
\right).
\end{equation}

To control memory consumption, the implementation evaluates at most 50,000
activation values for each layer using deterministic strided subsampling.

The activation damage coefficient is therefore

\begin{equation}
L_{n,a}^{A}
=
\mathcal{F}_n^A
\epsilon_{n,a}^{A}.
\end{equation}

The aggregate activation contribution is

\begin{equation}
L_A
=
\sum_n
\sum_{a\in\mathcal{C}^{A}}
L_{n,a}^{A}q_{n,a}^{A}.
\end{equation}

\subsection{Asymmetric Activation Search Space}

Preliminary joint weight--activation experiments initially allowed lower
activation precisions. These configurations resulted in severe degradation,
motivating a restricted activation search space for the final formulation:

\begin{equation}
\mathcal{C}^{W}
=
\{4,5,6,7,8\},
\qquad
\mathcal{C}^{A}
=
\{6,7,8\}.
\end{equation}

The preliminary local-gradient experiments motivated changing both the
activation profile and candidate menu (Section~\ref{sec:development_main}).
They are not a controlled menu-only ablation of this empirical formulation.

\subsection{Logarithmic Penalty Sweeping}

Because stochastic QUBO optimization can produce non-monotonic changes in
the selected architecture as the compute penalty changes, the main Pareto
analysis samples the global penalty coefficient $\gamma$ over a logarithmic
range.

For
$[\gamma_{\min},\gamma_{\max}]$, the sampled values are

\begin{equation}
\gamma_i
=
10^{
\log_{10}(\gamma_{\min})
+
i
\frac{
\log_{10}(\gamma_{\max})
-
\log_{10}(\gamma_{\min})
}{
N-1
}
},
\qquad
i\in\{0,\ldots,N-1\}.
\end{equation}

Development exploratory sweeps use 500 simulated-annealing reads and the
preliminary refinement examples use 1,000 reads for their seed. The expanded
target-BOP comparisons use a different procedure: gamma search with five
reads per step, at most 20 expansion steps and 14 binary-search steps, followed
by an independent 500-read solve. Requested target BOPs therefore need not
equal the achieved discrete cost.

% ============================================================================
% 4. TOPOLOGICAL CONSTRAINTS
% ============================================================================

\section{Structural Priors and Solver Details}

The QUBO objective incorporates interactions between variables associated with
different parts of the computational graph. These terms are implemented as
quadratic off-diagonal contributions.

\subsection{BOPs-Style Weight--Activation Coupling}

Let the summation index range over the selected, routed convolutions. Fixed
full-precision layers and non-convolutional operators are excluded from this
proxy. The computational cost of a convolutional layer is modeled using the
BOPs-style proxy

\begin{equation}
C_{\mathrm{BOPs}}
=
\frac{
\sum_n
MAC_n b_n^W b_n^A
}{
\sum_m
MAC_m32^2
}.
\end{equation}

After substituting the One-Hot representation,

\begin{equation}
C_{\mathrm{BOPs}}
=
\sum_n
\sum_{b\in\mathcal{C}^{W}}
\sum_{a\in\mathcal{C}^{A}}
\left(
\frac{
MAC_nba
}{
\sum_m MAC_m32^2
}
\right)
q_{n,b}^{W}q_{n,a}^{A}.
\end{equation}

The resulting interaction is multiplied by the global compute penalty
$\gamma$ in the QUBO matrix.

\subsection{Skip-Connection Compatibility}

The implemented model contains the three-branch fusion

\begin{equation}
X_{\mathrm{fused}}
=
x_1+u_2+u_3.
\end{equation}

The corresponding activation variables are grouped as

\begin{equation}
\mathcal{A}_{\mathrm{skip}}
=
\{
n_{\mathrm{res}},
n_{\mathrm{up},2},
n_{\mathrm{up},3}
\}.
\end{equation}

Here $n_{\mathrm{res}}$ identifies a residual-branch convolution and
$n_{\mathrm{up},2}$ and $n_{\mathrm{up},3}$ identify two upsampling-path
convolutions selected as fusion-path proxies. For two activation states with different bit-widths,
$a_1\neq a_2$, the assembled QUBO receives an exclusion penalty:

\begin{equation}
Q_{(i,a_1),(j,a_2)} \leftarrow Q_{(i,a_1),(j,a_2)} + 2\alpha.
\end{equation}

This encourages compatible activation bit-widths among the participating
branches. Equal bit-widths do not by themselves imply identical scales,
offsets, or dynamic ranges; the constraint is therefore interpreted as a
structural compatibility prior rather than a guarantee of identical dynamic
ranges or an integer-only addition interface. The activation variables refer
to convolution inputs during evaluation; these positions proxy the fusion
paths rather than quantizing all three tensors entering the addition.

\subsection{Module-Order Activation Interactions}

The cross-layer interaction implemented by the proposed QUBO acts on
activation variables. The implemented pairs are consecutive convolutions in module traversal order,
which is not a traced data-flow graph and can cross parallel paths. For a
selected ordered pair $n$ and $n+1$, the interaction
between candidate activation states $a_i$ and $a_j$ is

\begin{equation}
Q_{(n,a_i),(n+1,a_j)} \leftarrow Q_{(n,a_i),(n+1,a_j)} + \omega L_{n,a_i}^{A} L_{n+1,a_j}^{A},
\end{equation}

where

\begin{equation}
L_{n,a}^{A}
=
\mathcal{F}_n^A
\epsilon_{n,a}^{A}.
\end{equation}

The reported advanced formulation uses

\begin{equation}
\omega=0.15.
\end{equation}

This interaction penalizes simultaneous large estimated activation damage
at paired positions in that ordering. We interpret it as an ordered-layer
regularizer, rather than an exact model of error propagation along graph edges.

\subsection{Unified QUBO Matrix}

Before assembling the QUBO matrix, the diagonal loss coefficients are
normalized by their maximum value:

\begin{equation}
L_{\max}
=
\max_u L_u,
\end{equation}

\begin{equation}
\widehat{L}_u
=
\frac{L_u}{L_{\max}}.
\end{equation}

Because the adjacent-layer penalty is formed by a product of two loss
coefficients, its implementation is normalized by

\begin{equation}
\widehat{L}_{ij}^{\mathrm{cross}}
=
\frac{
L_{ij}^{\mathrm{cross}}
}{
L_{\max}^2
}.
\end{equation}

The largest BOP coupling coefficient is denoted by

\begin{equation}
C_{\max}^{\mathrm{BOP}}
=
\max_{(i,j)}
C_{ij}^{\mathrm{BOP}},
\end{equation}

and the largest normalized cross-layer loss coefficient by

\begin{equation}
L_{\max}^{\mathrm{cross}}
=
\max_{(i,j)}
\widehat{L}_{ij}^{\mathrm{cross}}.
\end{equation}

For $\beta=1$ in the reported implementation, the exclusion penalty is
adapted as

\begin{equation}
\alpha
=
2
\left[
\beta
+
\beta L_{\max}^{\mathrm{cross}}
+
\gamma C_{\max}^{\mathrm{BOP}}
+
1
\right].
\end{equation}

Let $x_u\in\{0,1\}$ represent a precision-assignment state. The QUBO energy
is

\begin{equation}
E(x)
=
\sum_u Q_{u,u}x_u
+
\sum_{u<v}Q_{u,v}x_ux_v.
\end{equation}

The implemented diagonal coefficients are

\begin{equation}
Q_{u,u}
=
\beta\widehat{L}_u-\alpha.
\end{equation}

For conflicting one-hot states,

\begin{equation}
Q_{u,v}
\mathrel{+}=2\alpha.
\end{equation}

For weight--activation BOP coupling,

\begin{equation}
Q_{u,v}
\mathrel{+}
=
\gamma
C_{u,v}^{\mathrm{BOP}}.
\end{equation}

For adjacent-activation interactions,

\begin{equation}
Q_{u,v}
\mathrel{+}
=
\beta
\widehat{L}_{u,v}^{\mathrm{cross}}.
\end{equation}

The skip-mismatch pairs also receive $2\alpha$, as described above; multiple
contributions to the same pair are added. All remaining entries are zero.

This explicit matrix construction corresponds to the numerical QUBO assembled
in the implementation.

\subsection{Simulated Annealing Search}

The QUBO matrix is optimized using classical simulated annealing.
For the primary experiments, the solver evaluates the requested number of
annealing reads with random seed 123. The seed is exposed explicitly so that
stochastic robustness can be assessed in a separate multi-seed experiment.
For each run, the configuration with the lowest observed QUBO energy is
selected:

\begin{equation}
x^{*}
=
\arg\min_{x\in\mathcal{X}_{\mathrm{observed}}}
E(x).
\end{equation}

Because Simulated Annealing is a stochastic heuristic and the search is not
exhaustive, $x^*$ is treated as the \emph{best observed solution under the
specified solver budget}, rather than as a certified global optimum.
The extended solver then repairs invalid one-hot groups to their highest
precision state and performs up to four coordinate-improvement passes over
layer groups. This postprocessing is separate from raw annealing; reported
raw solver energy need not equal the energy of a subsequently repaired route.
The corrected all-convolution seed study uses its separate raw-solver path.

% ============================================================================
% 5. PROTES
% ============================================================================

\section{Architecture and PROTES Implementation}

The QUBO stage evaluates candidate precision assignments using a surrogate
objective combining sensitivity-weighted quantization damage and
computational cost. PROTES is subsequently used to refine the best observed
QUBO solution using direct network evaluation.

\subsection{Model Architecture}

The experimental HalfUNet model is a custom compact encoder--decoder. The architecture contains
NAFBlock-based residual units, depthwise convolution, pointwise convolution,
channel-wise layer normalization, multiplicative gating, simplified channel
attention, and pixel-shuffle
upsampling. Its final representation is obtained using the three-way feature
fusion described previously.

The NAFBlock design is based on NAFNet~\cite{chen2022simple}. However, the
overall network used in this work is a custom architecture and is not the
canonical NAFNet or the published Half-UNet architecture.

\subsection{PROTES Search Space and QUBO Seeding}

For a model containing $L$ optimized layers, PROTES operates on a discrete
index vector of dimension

\begin{equation}
d=2L,
\end{equation}

where the first $L$ coordinates represent weight precision states and the
remaining $L$ coordinates represent activation precision states.

The best observed QUBO solution is converted to the corresponding PROTES
index vector $i_\star$. This vector is used to construct a peaked initial TT
probability distribution. The same solution is also supplied through the
seed-pool mechanism during the first five PROTES iterations.

The resulting procedure is a seeded probabilistic search; it does not impose
a hard Hamming-distance constraint around the QUBO solution.

\subsection{ROI-Constrained PROTES Objective}
\label{sec:roi_objective}

Let $B$ be the candidate's normalized routed-layer BOP ratio and
$B_0=B(\mathcal A_{\mathrm{QUBO}})$ the seed ratio. The implemented
restoration objective is the continuous piecewise score
\begin{equation}
S(\mathcal A)=\operatorname{PSNR}(\mathcal A)-\gamma B
 -(h-\gamma)\max(B-B_0,0),\qquad h=\max(10\gamma,100).
\label{eq:roi}
\end{equation}
Thus, for $B\leq B_0$ the cost is $\gamma B$, while above the seed
it is $\gamma B_0+h(B-B_0)$. Only the excess receives the larger
slope; there is no discontinuous charge on the whole candidate cost.
At $\gamma=1.584893$, increasing $B$ by 0.01 above the seed incurs
an additional 1.0 dB penalty. This is a soft penalty, not a hard compute
constraint, so selected candidates can exceed the seed cost. The QUBO
coefficient and the black-box penalty are reused numerically in these runs,
but operate on different objective scales. For classification, top-1
accuracy in percentage points replaces PSNR in the direct-evaluation score.

\subsection{PROTES Evaluation Strategy}

For the repeated-search reviewer experiment, PROTES uses sampling batch size
50, ten elites, tensor-train rank five and learning rate $0.05$. The seeded
initialization uses concentration eight and exploration probability $0.03$;
30\% seed-pool sampling is enabled for five iterations with five random
samples. Search seeds are 1701, 1702 and 1703, and recovery uses seed 77003
for the selected target. The five validation-subset runs keep optimizer seed
4402 fixed, with subset-index seed 4401 and recovery seed 88003.

The optimizer seed of the preliminary W+A sweep is unverified. The
repeated-search experiment explicitly controls the optimizer seeds listed
above; these studies therefore provide different evidence about stochastic
variation.

During optimization, each candidate architecture is evaluated using a
32-patch validation subset to reduce the cost of repeated network evaluation.
The best candidate found by PROTES is subsequently evaluated on the complete
test loader.

The reported PROTES configuration uses a batch size of 50 candidates, keeps
the 10 highest-scoring candidates as elites, uses TT rank 5, an Adam learning
rate of $5\times10^{-2}$, and peaked
initialization around the QUBO solution, a seed-pool fraction of 0.3 during
the first five iterations, and an exploration probability of 0.03.
The primary full-sweep cost records report a nominal budget of 5,000;
the matched-method, subset-robustness and generalization experiments use
2,000. Development refinement examples report their own 1,000/2,000 budgets.
The primary search pads the activation alphabet to match the five weight choices
and copies the first activation choice in each coupled group to the others;
thus different sampled indices may decode to the same route. Cache counts
in the original sweep count sampled index vectors, whereas the repeated-
search study counts canonical assignments.

% ============================================================================
% 6. EXPERIMENTAL SETUP AND RESULTS
% ============================================================================

\section{Full Experimental Record}\label{app:experiments}

\begin{table}[tbp]
\centering
\small
\caption{Experiment variants and evidence scope. Costs use the routed-layer
FP32 denominator; static and recovered results are reported separately.}
\label{tab:protocol_variants}
\begin{tabular}{p{0.25\linewidth}p{0.29\linewidth}p{0.35\linewidth}}
\toprule
Record & Routing and split & Scope \\
\midrule
Primary comparison & 38 convolutions / 304 variables; patch-random SIDD & Single original recovery evaluations; repeated routes at loose targets \\
Repeated search and subset studies & Protected primary routing; 32-patch validation inner loop & Search-seed/subset variation conditional on fixed recovery settings \\
Corrected paired statistics and solver robustness & 40 convolutions / 320 variables; patch-random SIDD & Static paired patch tests and solver-seed variation \\
Supporting ablations & All-convolution variant & Exploratory supporting ablations; not recovered main-result significance \\
Additional architectures & Model-specific routed convolutions; source-image-disjoint SIDD or CIFAR-10 & Separate dense checkpoints; shared physical scenes in SIDD; one recovery replicate \\
Formulation development & Earlier routing, menus and optimizer budgets & Contextual observations; not pooled with the primary comparison \\
\bottomrule
\end{tabular}
\end{table}

We evaluate the framework through a sequence of experiments that isolate the
effects of task-aware sensitivity, activation handling, topological
constraints, and PROTES refinement.

\subsection{Implementation Details and Metrics}

The repeated-search and additional-architecture runs use
Linux, Python 3.10.20, PyTorch 2.9.1+cu128, CUDA 12.8 and an NVIDIA
H100 80GB HBM3 GPU. The repeated-search environment uses cuDNN 9.2.
These settings identify the added runs; they are not retroactively assigned
to preliminary statistical or full-sweep runs. Reviewer search seeds are
1701--1703; its recovery seed is $77000+t$, where $t$ is the zero-based
target index. The subset study uses seeds 4401 (selection), 4402 (search)
and 88000 (recovery). Additional-architecture recovery uses base seed 42
with target offset $1000t$.

The primary architecture is the custom NAFBlock-based HalfUNet
implementation described above. It accepts three-channel inputs and uses a
base channel width of 32. The network contains two encoder downsampling
stages, a central NAFBlock stage, two pixel-shuffle upsampling paths, and a
three-branch feature-fusion operation.

The primary experiments use 160 noisy/ground-truth pairs from SIDD Small
sRGB~\cite{abdelhamed2018high}. Each image is divided into four quadrants,
which are resized to $256\times256$ RGB patches. The resulting 640 pairs
are randomly split into 512 training, 64 validation and 64 test patches
using seed 42. Each training patch is accessed three times, with an unaugmented
access and stochastic flip augmentations, giving 1,536 training items.
Validation and test patches are not augmented; the primary batch size is 16.

This split occurs after patch extraction. Different patches from the same
source image can therefore appear in different partitions. We call it the
\emph{patch-random} protocol; these resized-patch scores are not official
full-resolution SIDD benchmark results.

The expanded mixed-precision comparison protects the first and last
convolutions, routing 38 internal convolutions (304 binary variables).
The paired statistics, corrected SA-seed study, omega/beta, probe-bit,
activation-menu and bias studies instead route all 40 convolutions
(320 variables). We report these as an \emph{all-convolution variant},
not as repetitions of the protected-layer main comparison.

For final architecture evaluation, the baseline network is deep-copied and
the selected weight bit-width is applied independently to each selected
convolutional layer. Development weight quantization uses the symmetric minimum-MSE scale
search described above with an 80-point scale grid; expanded evaluations
use 64 points as specified in Appendix~\ref{app:formulation}. Convolutional biases are
not quantized in the reported experiments and therefore remain at their
original precision. Activation quantization is applied dynamically during
forward evaluation using the selected activation bit-width.

The computational cost is represented by a normalized BOPs proxy. For layer
$n$,

\begin{equation}
BOPs_n
=
MACs_n
b_{W,n}
b_{A,n}.
\end{equation}

The Compute Cost Ratio is

\begin{equation}
\operatorname{Compute\ Cost\ Ratio}
=
\frac{
\sum_n
MACs_n b_{W,n}b_{A,n}
}{
\sum_n
MACs_n32^2
}.
\end{equation}

This quantity is a computational proxy based on the assigned precision and
MAC count; it is not intended to represent measured hardware latency.

Network restoration quality is evaluated using PSNR and SSIM
~\cite{wang2004image}, with RGB outputs clipped to $[0,1]$ and data range one.
The shared restoration evaluator computes PSNR from the MSE over each batch
and averages the batch scores. The paired analysis separately averages
per-patch PSNR, which is a different aggregation. Dynamic symmetric
activation quantization uses the maximum magnitude over the current tensor,
including its batch dimension; batch size is therefore part of the protocol.
Average weight bits are parameter-count weighted, average activation bits
are MAC weighted, and selected-weight storage reduction is
$1-\sum_n |W_n|b_n^W/(32\sum_n|W_n|)$. Development layer-average bit counts
are unweighted unless otherwise specified.

The static implementation retains floating-point weights after rounding and
uses floating-point convolution, bias, accumulation, normalization, residual
addition and gating arithmetic. The extended LSQ+ protocol uses unsigned
asymmetric activation quantizers, four epochs updating quantizer parameters
and fourteen epochs also updating convolution parameters, with a cap of 100
batches per epoch. Adam step sizes are $10^{-4}$ for scales/offsets and
$10^{-5}$ for convolution parameters; scale/offset rates are halved in the
second stage. Restoration recovery uses L1 loss; classification uses
cross-entropy. This recovery is distinct from the MSE-based preliminary
sensitivity calculation and from isolated scale calibration.

The profiling caps are 10 validation batches for weight gradients and
candidate activation error and five for the global activation probe. With
64 validation patches at batch size 16, the primary loader contains only
four batches, so these caps do not imply ten distinct processed batches.
All sensitivity/calibration data come from validation, not the test split.

\subsection{Formulation Development: Full Method Chronology}
\label{sec:development_record}

The preliminary development records explain two distinct revisions: the
weight-damage objective and the activation-sensitivity estimator. They are
before-recovery evaluations from earlier protocols, not controlled ablations
of the final protected-layer comparison.

\paragraph{Weight-only development.}
The original bit-reduction objective produced an imbalanced route: 35.778 dB
at a reported average of 7.94 weight bits. One-hot selection using
parameter-count-weighted reconstruction MSE reached 36.803 dB near six bits.
Replacing that damage proxy with normalized squared-gradient sensitivity and
removing the additional parameter-count multiplier reached 36.925 dB
(Table~\ref{tab:weight_progression}). The 0.122 dB difference between the two
one-hot routes is the recorded incremental gain of task-aware weighting;
relative to uniform six-bit weights, the gain is 0.298 dB. These preliminary
averages do not establish measured storage savings or solver optimality.

\paragraph{Joint-allocation phases.}
At $\gamma=20$, Phase 1 reused the weight-gradient score for both weights
and activations. Phase 2, also reported at $\gamma=20$, instead collected squared activation gradients during backpropagation and used separate local scores for the two tensors. It yielded
25.481 dB at reported averages of 6.10 weight bits and 6.00 activation bits.
Phase 3 multiplied the activation-damage term by $\lambda_A=50$; its
$\gamma=5$ run recorded
24.896 dB shows that scalar reweighting did not resolve the failure at that
operating point. A temporary fallback retained QUBO weight routing and fixed
activations at eight bits, reaching 36.448 dB. These are different estimators
and objectives, not simply an unrestricted-activation variant of the final
empirical method.

The later method perturbed one intermediate activation with temporary
four-bit quantization and measured the resulting reconstruction MSE at the
network output across validation data. It used that downstream response to
construct a global empirical activation-sensitivity score, alongside candidate
activation reconstruction errors. It also restricted the activation menu to
$\{6,7,8\}$ while allowing weights in $\{4,5,6,7,8\}$. Its reported
$\gamma=50$ route used 5.58-bit weights and 7.17-bit activations and achieved
35.105 dB at 3.603\% BOPs (Table~\ref{tab:joint_progression}). This activation
average is consistent with the final menu. Profiling and the menu changed
together, so the preliminary phase comparison cannot isolate their separate
contributions or prove an intrinsic six-bit activation threshold. The later
menu-only study appears separately in Appendix~\ref{sec:activation_cliff}.

\subsection{Ablation Study: Structural Priors}
\label{sec:topological_ablation}

We compare the explored quality--cost operating points for three formulations:

\begin{enumerate}
    \item \textbf{Naive QUBO (No Skip, No Cross):}
    the task-aware formulation without graph-aware interactions.

    \item \textbf{QUBO + Skip Constraints:}
    the task-aware formulation with the skip-connection compatibility
    penalties.

    \item \textbf{Full (Skip + Cross-Layer):}
    the formulation additionally including the adjacent-activation
    interaction with $\omega=0.15$.
\end{enumerate}

\begin{figure}[tbp]
\centering
\includegraphics[width=0.9\columnwidth]{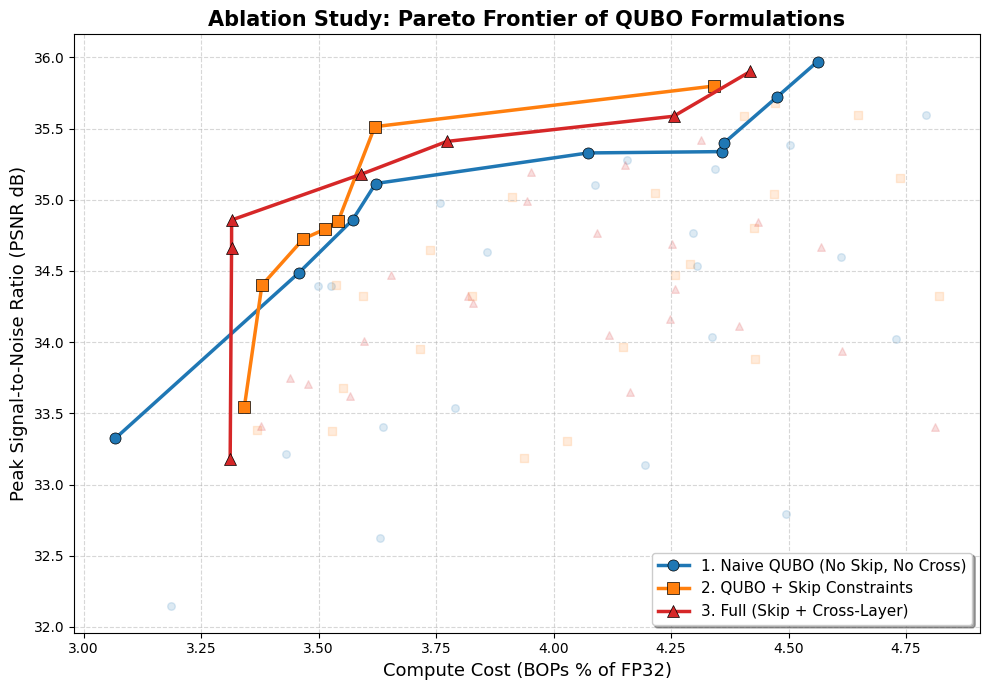}
\caption{
Ablation study of QUBO formulations. Scattered points represent candidate
architectures evaluated during the stochastic search, while the solid curves
connect reported operating points and are not a certified Pareto frontier. Adding skip-connection and
adjacent-activation interactions improves the observed PSNR--compute
frontier over the tested compression regime.
}
\label{fig:qubo_ablation}
\end{figure}

The Naive QUBO formulation exhibits pronounced performance degradation below
approximately 3.5\% BOPs. The Skip-Constrained formulation improves the
observed frontier in the intermediate compression regime, while the Full
formulation provides additional recovery at more aggressive compression
levels.

The Full formulation reaches approximately 35.9 dB at 4.4\% BOPs in the
reported sweep. These results support the use of selected structural priors
within the evaluated QUBO formulation, but they do not establish global
optimality over the complete search space.

\subsection{Development Study: QUBO-Seeded PROTES}
\label{sec:progression_protes}

The best observed QUBO configuration is used to initialize PROTES. For the
reported end-to-end refinement experiment, the QUBO seed is obtained with
1000 Simulated Annealing reads.

The report describes an expansion tax with rate $\max(10\gamma,100)$
and evaluates refinement on 32 validation patches. These preliminary records
are separate from the expanded matched-method comparison; the report
description does not establish exact equivalence to the continuous
refinement score in Eq.~\ref{eq:roi}. The linear-score comparison appears
in main Table~\ref{tab:penalty_main}.

\begin{table}[tbp]
\centering
\caption{
ROI-Constrained PROTES Refinement across Selected $\gamma$ Regimes.
}
\label{tab:protes_ablation}
\resizebox{\columnwidth}{!}{
\begin{tabular}{lccccc}
\toprule
\textbf{Target $\gamma$}
&
\textbf{Budget ($m$)}
&
\textbf{QUBO Seed}
&
\textbf{PROTES Final}
&
\textbf{$\Delta$ BOPs}
&
\textbf{$\Delta$ PSNR}
\\
\midrule
$\gamma=85.77$
&
1000
&
3.323\% / 35.144 dB
&
3.308\% / 35.693 dB
&
-0.016\%
&
+0.549 dB
\\

$\gamma=85.77$
&
2000
&
3.323\% / 35.144 dB
&
3.165\% / 35.704 dB
&
-0.158\%
&
+0.560 dB
\\

$\gamma=11.66$
&
2000
&
4.119\% / 35.969 dB
&
4.042\% / 36.300 dB
&
-0.078\%
&
+0.331 dB
\\

$\gamma=1.58$
&
2000
&
3.990\% / 35.497 dB
&
3.985\% / 36.318 dB
&
-0.004\%
&
+0.821 dB
\\
\bottomrule
\end{tabular}
}
\end{table}

The comparison at $\gamma=85.77$ uses two different PROTES evaluation
budgets, $m=1000$ and $m=2000$, while keeping the QUBO seed fixed. These
are single stochastic runs with different budgets, not a repeated controlled
estimate of the budget effect.

\subsection{Main Mixed-Precision Comparison}
\label{sec:main_comparison}

The HAWQ-style baseline combines normalized absolute Hutchinson weight-Hessian
trace estimates with calibrated weight reconstruction error and the shared
activation-error/sensitivity term. It uses two one-sample batches and two
probes per batch (seed 2026), followed by a multiple-choice dynamic-programming
allocator with resource resolution 100,000. This is a sensitivity--allocation
adaptation, not the complete HAWQ-V3 deployment system.

Main Table~\ref{tab:main_comparison} reports the
original six-target records, reporting achieved rather than requested cost.
Static/LSQ+ rows are paired by assignment ID; reused routes share their static
evaluation. Later additional uniform recovery runs and dense gamma-sweep
points are not pooled into these single-run rows. These are descriptive
operating points, not seed-averaged estimates or equal-cost comparisons.

At the 4.102\% requested target, QUBO + PROTES achieves 37.192 dB after
LSQ+ at 4.035\% BOPs, compared with 37.132 dB at 4.149\% for QUBO
and 37.092 dB at 4.101\% for HAWQ-style allocation. The original uniform
row gives 36.802 dB; an additional recovery run on that same uniform
assignment gives 37.010 dB, illustrating variation not captured by a single
row. At the lowest target, QUBO alone reaches 36.997 dB after LSQ+,
exceeding the refined route's 36.823 dB despite the latter's stronger static
result. Refinement gains depend on both the operating point and recovery.

The three loosest requested targets reuse the same QUBO assignment and the
same refined assignment within each method. Their differing LSQ+ scores
reflect separate recovery runs, not three distinct routing solutions. No
EdMIPS or ALPS/EAGL comparison is included in the completed experiments.

Weight-storage compression is reported numerically at the representative
operating point used for the deployment analysis.  Because the current
implementation keeps floating-point tensors in the serialized model state,
we distinguish the analytical packed-storage reduction from the realized
serialized model size.

\begin{table}[t]
\centering
\caption{Representative computational and storage characteristics at the
executed deployment operating point. BOPs and ideal packed storage are
analytical quantities; the current reference implementation does not pack
weights into arbitrary-bit tensors.}
\label{tab:storage_deployment}
\resizebox{\columnwidth}{!}{%
\begin{tabular}{lccc}
\toprule
\textbf{Method} & \textbf{BOPs (\%)} & \textbf{Whole-model reduction (\%)} &
\textbf{Ideal packed size (KiB)} \\
\midrule
Uniform target-matched & 4.102 & 77.592 & 61.31 \\
HAWQ-style Hessian + knapsack & 4.101 & 78.486 & 58.87 \\
Enhanced QUBO & 4.149 & 79.745 & 55.43 \\
Enhanced QUBO + PROTES & 4.035 & 79.080 & 57.25 \\
\bottomrule
\end{tabular}}
\end{table}

\subsection{Paired Test-Set Comparison: Actual QUBO Routing vs. Uniform W4/A8}
\label{sec:paired_qubo_uniform}

We additionally perform a paired patch-level comparison between the actual
QUBO routing produced by the solver and the uniform W4/A8 baseline. The QUBO
configuration is decoded from the lowest-energy solution obtained with 500
Simulated Annealing reads at $\gamma=200$, using the primary solver seed
123 and the reported structural settings. Both architectures are evaluated
on the same 64 held-out test patches, producing one PSNR value per patch before
computing paired differences.

The actual QUBO routing uses 3.2873\% BOPs, compared with 3.1250\% for
Uniform W4/A8. Its mean per-patch PSNR is 34.0338$\pm$1.5691 dB versus
33.6569$\pm$1.4435 dB for the uniform baseline, corresponding to a mean
paired improvement of 0.3769 dB. The 95\% confidence interval for the mean
paired difference is [0.1748, 0.5790] dB. A paired $t$-test gives
$t=3.7269$ with $p=4.17\times10^{-4}$, while the Wilcoxon signed-rank test
gives $p=1.10\times10^{-3}$. The paired Cohen's $d_z$ is 0.466, and the
QUBO routing has higher PSNR on 42 of the 64 test patches (65.6\%).

This comparison supports a statistically detectable quality improvement for
the evaluated QUBO assignment, but the QUBO assignment also uses slightly
more BOPs than the uniform W4/A8 baseline. We therefore interpret this result
as a paired quality comparison at a modestly higher analytical compute cost,
rather than as evidence of dominance at matched BOPs. The displayed
standard deviations describe patch variation, not solver-seed variation.
These tests treat patch differences as independent; shared source
images/scenes and batch-dependent activation scales limit that assumption.
The intervals and p-values are therefore nominal patch-level results,
not a scene-level significance claim or a final LSQ+ comparison against
HAWQ-style routing.

\begin{table}[t]
\centering
\caption{Paired test-set comparison between the actual QUBO routing and
uniform W4/A8 on the same 64 held-out patches.}
\label{tab:paired_qubo_uniform}
\begin{tabular}{lcc}
\toprule
\textbf{Metric} & \textbf{Actual QUBO} & \textbf{Uniform W4/A8} \\
\midrule
BOPs (\%) & 3.2873 & 3.1250 \\
Mean PSNR (dB) & 34.0338$\pm$1.5691 & 33.6569$\pm$1.4435 \\
\midrule
Mean paired $\Delta$PSNR (dB) & \multicolumn{2}{c}{+0.3769} \\
95\% CI for paired $\Delta$ (dB) & \multicolumn{2}{c}{[0.1748, 0.5790]} \\
Paired $t$-test $p$ & \multicolumn{2}{c}{$4.17\times10^{-4}$} \\
Wilcoxon signed-rank $p$ & \multicolumn{2}{c}{$1.10\times10^{-3}$} \\
Cohen's $d_z$ & \multicolumn{2}{c}{0.466} \\
QUBO better / Uniform better & \multicolumn{2}{c}{42 / 22} \\
\bottomrule
\end{tabular}
\end{table}

\subsection{Search-Method and Initialization Analysis}
\label{sec:search_ablation}

At a requested BOP ratio of $4.1015625\%$, we compare four stochastic search
procedures using seeds 1701, 1702, and 1703 and a nominal budget of 2,000
proposals per run. All use the same 32-patch validation subset. We also
evaluate the fixed QUBO assignment used to initialize refinement. Actual
proposal counts, unique network evaluations, and achieved BOP ratios differ
between methods and are reported explicitly.

Quality and achieved costs appear in main Table~\ref{tab:search_main};
we give proposal counts and runtimes here.

Random search uses 2,000 proposals and 2,000 unique evaluations on average.
Random PROTES uses 2,001 proposals and 2,000 unique evaluations; QUBO-seeded
PROTES uses 2,001 proposals and approximately 1,952.7 unique evaluations.
Simulated annealing uses 2,006 proposals and 1,700 unique evaluations. The
fixed QUBO assignment is evaluated once. These counts distinguish nominal
search budgets from realized black-box evaluation costs.

QUBO-seeded PROTES achieves the highest mean static PSNR among the tested
stochastic procedures. Relative to the fixed QUBO route, its mean improvement
is approximately 0.046 dB before LSQ+ and 0.007 dB after LSQ+. These are
descriptive differences at slightly different achieved BOP ratios. They
support a limited benefit of refinement at this operating point, but do not
establish general superiority over QUBO alone or the other search procedures.

The mean search runtimes are 1,206.6 s for random search, 1,262.6 s for
random PROTES, 1,058.2 s for black-box SA and 1,249.8 s for seeded PROTES.
These exclude final LSQ+ recovery. The QUBO-alone timer only evaluates an
already selected route; profiling and solving are reported separately in
Section~\ref{sec:search_cost}. Black-box SA searches on the network score
and is distinct from annealing the QUBO surrogate.

\subsection{Hardware and Deployment Analysis}
\label{sec:hardware}

Because normalized BOPs are a computational proxy, we additionally measured
wall-clock inference latency for representative assignments on the available
CPU and GPU hardware. The benchmark uses the same input resolution and model
implementation as the primary evaluation. The current reference implementation
performs fake quantization and then executes floating-point convolution
kernels; it therefore does not provide native arbitrary-bit arithmetic.

This distinction is important when interpreting the BOP results. The BOP ratio
and storage-compression values describe the assigned numerical precision, while
measured latency reflects the implementation used to execute those assignments.
Consequently, a reduction in analytical BOPs is not expected to yield a
proportional latency reduction until native low-bit kernels, packed memory
layouts, and suitable hardware support are introduced. We therefore do not use
the present latency measurements as evidence of hardware acceleration; instead,
they document the deployment behavior of the reference implementation.

The benchmark uses batch size one, $3\times256\times256$ inputs,
25 warmups and 75 CPU / 200 GPU timed repetitions. Table~\ref{tab:hardware_latency}
reports means and standard deviations. Speedup is FP32 latency divided by
method latency; values below one indicate slower execution. CPU timing
variance is substantial, so small differences between routes should not be
overinterpreted. The recorded GPU is an NVIDIA H100 80GB HBM3;
the CPU model and thread configuration are not recorded. These timings
should not be extrapolated to other platforms.

Serialized model states remain 322,499 bytes on CPU and 324,227 bytes
on GPU for every method. Peak GPU allocation is 122,115,072 bytes for FP32
and 129,980,416 bytes for each quantized variant. Neither measured
serialization nor peak allocation improves. No energy value was recorded.
The deployment storage column describes ideal whole-model packed reduction,
unlike the selected-weight reduction in routing summaries (for uniform
W6/A7 these are 77.592\% and 81.250\%, respectively).

\begin{table}[t]
\centering
\caption{Measured inference latency at the representative operating point. Latency is measured on the reference PyTorch
implementation with fake quantization followed by floating-point kernels.}
\label{tab:hardware_latency}
\small
\begin{tabular}{lccc}
\toprule
Method & CPU (ms) & GPU (ms) & FP32 / GPU latency \\
\midrule
FP32 & $120.36\pm67.61$ & $3.002\pm0.094$ & 1.000 \\
Uniform & $188.71\pm135.86$ & $6.776\pm0.239$ & 0.443 \\
HAWQ-style & $169.85\pm102.63$ & $6.743\pm0.196$ & 0.445 \\
QUBO & $162.50\pm77.66$ & $6.735\pm0.249$ & 0.446 \\
QUBO + PROTES & $179.27\pm100.54$ & $6.769\pm0.214$ & 0.444 \\
\bottomrule
\end{tabular}
\end{table}

\subsection{PROTES Validation-Subset Robustness}
\label{sec:protes_subset}

Because PROTES repeatedly evaluates candidate networks using a small validation
subset, we tested whether the refinement process is sensitive to the particular
images selected for this inner loop. Five distinct 32-patch validation
subsets were evaluated while keeping the QUBO initialization, optimizer seed,
target operating point, candidate space, and evaluation budget fixed. Each
refined assignment was subsequently evaluated on the complete held-out test
set.

Across the five subsets, the resulting static-test PSNR was
$36.712\pm0.037$ dB, indicating relatively small variation in final
restoration quality despite changes in the inner-loop validation sample.
However, the exact routing assignments were less stable: pairwise weight
agreement ranged approximately from 0.47 to 0.74, activation agreement from
0.68 to 0.92, and joint weight--activation agreement from 0.37 to 0.58.
Thus, the experiment supports stability of achieved restoration quality more
strongly than stability of the exact discrete architecture. The subsets overlap
(pairwise Jaccard indices 0.231--0.391). The requested target is 4.102\%,
subset-selection seed 4401 and optimizer seed 4402. After LSQ+, gains over
the QUBO reference range from $-0.034$ to $+0.033$ dB; the small static
gains do not consistently survive recovery.

\begin{table}[t]
\centering
\caption{Robustness of PROTES to the choice of the 32-patch inner-loop
validation subset. Agreement is computed pairwise across the five resulting
routing assignments.}
\label{tab:protes_subset_robustness}
\begin{tabular}{lc}
\toprule
\textbf{Quantity} & \textbf{Observed result} \\
\midrule
Static test PSNR & $36.712\pm0.037$ dB \\
Weight-bit agreement & 0.47--0.74 \\
Activation-bit agreement & 0.68--0.92 \\
Joint W+A agreement & 0.37--0.58 \\
\bottomrule
\end{tabular}
\end{table}

\subsection{Simulated Annealing Multi-Seed Robustness}
\label{sec:sa_multiseed}

Because the QUBO stage uses stochastic Simulated Annealing, we evaluated the
same formulation under five independent solver seeds (0--4) at three
representative compute penalties: $\gamma\in\{1.584893,11.659144,85.77\}$.
Each run used 500 annealing reads and the same profiling data and structural
hyperparameters. We report the mean and standard deviation of the resulting
test-set PSNR, BOP ratio, and QUBO energy across the five seeds.

The resulting PSNR means were 35.602$\pm$1.146 dB, 34.746$\pm$1.156 dB,
and 34.930$\pm$0.389 dB for $\gamma=1.584893$, 11.659144, and 85.77,
respectively. The corresponding BOP ratios were 4.109$\pm$0.366\%,
4.070$\pm$0.543\%, and 3.477$\pm$0.112\%. QUBO energy variation was
small within each penalty regime, with standard deviations of 0.043, 0.026,
and 0.086, respectively.

However, agreement between the exact discrete routings was substantially
lower than the stability of the objective values might suggest. Mean pairwise
coordinate agreement over the concatenated weight and activation vectors was 0.304, 0.281, and 0.311 across the
three penalty regimes. Weight-only agreement was 0.190, 0.163, and 0.238,
while activation-only agreement was 0.418, 0.400, and 0.385.
This coordinate agreement is the average of weight and activation agreement;
it differs from the both-precisions-at-one-layer agreement used in the subset
study. Thus, different annealing seeds can produce materially different discrete
routes while yielding close surrogate energies and, in two regimes, PSNR standard deviations exceeding 1 dB. This
result motivates reporting stochastic variation rather than treating a single
annealing realization as a uniquely identified architecture.

\begin{table}[t]
\centering
\caption{Five-seed Simulated Annealing robustness across three representative
QUBO compute penalties. Values are mean $\pm$ standard deviation over five
solver seeds.}
\label{tab:sa_multiseed}
\resizebox{\columnwidth}{!}{%
\begin{tabular}{lccc}
\toprule
$\gamma$ & Test PSNR (dB) & BOPs (\%) & QUBO energy \\
\midrule
1.584893 & $35.602\pm1.146$ & $4.109\pm0.366$ & $-321.972\pm0.043$ \\
11.659144 & $34.746\pm1.156$ & $4.070\pm0.543$ & $-331.907\pm0.026$ \\
85.77 & $34.930\pm0.389$ & $3.477\pm0.112$ & $-405.460\pm0.086$ \\
\bottomrule
\end{tabular}}
\end{table}

\subsection{Sensitivity to Topological and Objective Hyperparameters}
\label{sec:omega_beta}

The cross-layer interaction coefficient $\omega$ and QUBO objective scaling
coefficient $\beta$ control the relative contribution of structural
interactions and task-aware damage terms. We therefore evaluate
$\omega\in\{0,0.05,0.10,0.15,0.20,0.30,0.50\}$ and
$\beta\in\{0.5,1,2\}$ at two representative compression regimes,
$\gamma=11.66$ and $\gamma=85.77$.

All sweep quality values are evaluated on the test loader. The following
maxima are descriptive exploratory observations, not validation-selected
hyperparameters or independent evidence for choosing $\omega$.

The response is non-monotonic in both parameters. At $\gamma=11.66$, PSNR
ranges from 33.46 to 35.60 dB across the tested combinations, while BOPs range
from 3.66\% to 4.59\%. At $\gamma=85.77$, PSNR ranges from 32.89 to
35.76 dB and BOPs from 3.27\% to 4.09\%.

The best observed configuration in the sweep at $\gamma=11.66$ is
$(\omega,\beta)=(0.05,1)$, giving 35.604 dB at 4.210\% BOPs. At
$\gamma=85.77$, the best observed configuration is
$(\omega,\beta)=(0.15,2)$, giving 35.759 dB at 4.090\% BOPs. These
results show that the effect of cross-layer interaction depends on the
compression regime and objective scaling. We retain $\omega=0.15$ as
the preliminary setting, without treating this test-set sweep as a valid
selection procedure or evidence of a universal optimum.

\begin{table}[t]
\centering
\caption{Observed sensitivity to the cross-layer coefficient $\omega$ and
objective scaling coefficient $\beta$ at two representative compute
penalties. The reported configuration is an empirically selected setting,
not a claim of a universal optimum.}
\label{tab:omega_beta}
\begin{tabular}{lccc}
\toprule
$\boldsymbol{\gamma}$ & \textbf{PSNR range (dB)} & \textbf{BOPs range (\%)} &
\textbf{Best observed $(\omega,\beta)$} \\
\midrule
11.66 & 33.46--35.60 & 3.66--4.59 & (0.05, 1) \\
85.77 & 32.89--35.76 & 3.27--4.09 & (0.15, 2) \\
\bottomrule
\end{tabular}
\end{table}

\subsection{Robustness of Empirical Activation Sensitivity to Probe Precision}
\label{sec:probe_bits}

Empirical Global Activation Sensitivity uses a temporary quantization probe
to estimate the downstream effect of perturbing an intermediate activation.
To test whether the resulting sensitivity ranking depends strongly on that
probe precision, we repeated the profiling procedure using 3-, 4-, 5-, and
6-bit probes.

The pairwise Spearman correlations of the resulting layer rankings range from
0.936 to 0.984, indicating strong agreement in the relative ordering of layer
sensitivities. Pearson correlations are lower, ranging from 0.698 for the
3-bit versus 6-bit comparison to 0.967 for the 3-bit versus 4-bit comparison.
Thus, rankings are more stable than the absolute sensitivity scale.
Because coefficient magnitudes also enter the QUBO, rank agreement alone
does not establish identical routing or performance across probe precisions.

\begin{table}[t]
\centering
\caption{Pairwise correlations between activation-sensitivity rankings
obtained with different probe precisions.}
\label{tab:probe_bits}
\begin{tabular}{lcc}
\toprule
\textbf{Probe pair} & \textbf{Spearman} & \textbf{Pearson} \\
\midrule
3 vs. 4 bits & 0.978 & 0.967 \\
3 vs. 5 bits & 0.944 & 0.844 \\
3 vs. 6 bits & 0.936 & 0.698 \\
4 vs. 5 bits & 0.984 & 0.952 \\
4 vs. 6 bits & 0.969 & 0.852 \\
5 vs. 6 bits & 0.983 & 0.966 \\
\bottomrule
\end{tabular}
\end{table}

\subsection{Activation Search-Space Ablation}
\label{sec:activation_cliff}

The main search restricts activation candidates to $\{6,7,8\}$. To directly
test the motivation for this restriction, we compare it with an otherwise
identical search that allows $\{4,5,6,7,8\}$.

The unrestricted search exhibits a pronounced activation cliff. Its PSNR is
30.12 dB at $\gamma=1$, 26.10 dB at $\gamma=50$, and 27.68 dB at
$\gamma=100$. By contrast, the restricted activation search remains between
33.43 and 35.40 dB across the tested penalty values. These results show that
the tested QUBO search can select severely degraded assignments when lower
activation precisions are allowed. The requested unrestricted-menu PROTES
experiment has not been performed in this study, so these results do not
establish whether the failure is intrinsic to HalfUNet or reflects the
surrogate and search procedure.

\begin{table}[t]
\centering
\caption{Restricted versus unrestricted activation menus under QUBO search.}
\label{tab:activation_cliff}
\begin{tabular}{lccc}
\toprule
\textbf{Activation space} & \textbf{$\gamma$} & \textbf{BOPs (\%)} & \textbf{PSNR (dB)} \\
\midrule
$\{6,7,8\}$ & 1 & 4.482 & 35.126 \\
$\{6,7,8\}$ & 5 & 4.211 & 35.398 \\
$\{6,7,8\}$ & 50 & 3.237 & 33.426 \\
$\{4,5,6,7,8\}$ & 1 & 3.569 & 30.116 \\
$\{4,5,6,7,8\}$ & 50 & 3.527 & 26.101 \\
$\{4,5,6,7,8\}$ & 100 & 2.686 & 27.685 \\
\bottomrule
\end{tabular}
\end{table}

\subsection{Bias Precision Ablation}
\label{sec:bias_ablation}

The main experiments retain convolutional biases in their original precision.
To quantify the impact of this choice, we compare FP32 and INT8 bias
quantization for three representative uniform assignments. INT8 bias
quantization changes PSNR by $-0.145$ dB for Uniform W5/A7, $-0.036$ dB for
Uniform W6/A7, and $-0.128$ dB for W4/A8. The corresponding SSIM changes are
small.

These results indicate a modest quality penalty for the three tested uniform
configurations. They do not establish that bias precision leaves the relative
advantage of a mixed-precision assignment unchanged, and do not constitute
an integer-only deployment experiment. We therefore retain FP32 biases in the main experiments and
make this precision assumption explicit when interpreting the reported
quantization results.

\begin{table}[t]
\centering
\caption{Effect of INT8 bias quantization on representative assignments.}
\label{tab:bias_ablation}
\begin{tabular}{lcc}
\toprule
\textbf{Assignment} & $\Delta$PSNR (dB) & $\Delta$SSIM \\
\midrule
Uniform W5/A7 & -0.145 & -0.0018 \\
Uniform W6/A7 & -0.036 & -0.0004 \\
W4/A8 & -0.128 & +0.0001 \\
\bottomrule
\end{tabular}
\end{table}

\subsection{Computational Search Cost}
\label{sec:search_cost}

The protected-layer formulation contains 304 binary variables and 1,713
non-zero coefficients at the representative target. Its 512 exclusion pairs
comprise 494 within-layer one-hot conflicts and 18 skip-mismatch pairs.
There are also 570 W--A BOP couplings and 333 ordered-layer interaction
pairs; some keys overlap, so these counts are not disjoint. The
all-convolution variant has 320 variables and 1,803 coefficients.

\begin{table}[tbp]
\centering
\small
\caption{Offline costs at the 4.102\% primary requested target. Shared stages
are separated from target-specific search. Cache entries in the original
PROTES search count sampled index vectors, not necessarily unique routes.}
\label{tab:search_cost}
\begin{tabular}{lr}
\toprule
Stage & Time (s) \\
\midrule
Weight-gradient profiling & 0.387 \\
Candidate activation profiling & 0.233 \\
Global activation profiling & 4.834 \\
LSQ+ activation initialization & 20.488 \\
QUBO component construction & 5.074 \\
QUBO gamma search + independent 500-read solve & 1.819 \\
HAWQ-style Hessian profiling & 2.568 \\
HAWQ-style knapsack allocation & 0.059 \\
Original PROTES refinement & 529.075 \\
\bottomrule
\end{tabular}
\end{table}

The original refinement record has 4,890 cached index evaluations at this
target, approximately 0.108 s per entry, and a recorded nominal budget of
5,000. This differs from the 2,000-budget reviewer search ablation.
The recorded dense-checkpoint-to-final-LSQ+ times are 185.319 s for uniform,
205.565 s for HAWQ-style allocation, 198.140 s for QUBO and 734.474 s for
QUBO + PROTES. The corresponding LSQ+ training/materialization stages take
164.831, 177.986, 165.304 and 172.564 s. Totals include costs attributed
by the originating runs and exclude dense-model training; they should not
be reconstructed by summing every shared stage for every method.

Lower inference BOPs do not imply cheaper allocation. Refinement adds
hundreds of seconds here for a modest post-QAT improvement; QUBO alone or
Hessian--knapsack allocation remains a practical alternative. These timings
describe the recorded environment, not hardware-independent complexity.

\subsection{Generalization Across Data Splits, Architectures, and Tasks}
\label{sec:generalization}

Main Table~\ref{tab:generalization_main} reports the requested 4.1015625\%
operating point after LSQ+, with one recovery replicate (seed 42).

We additionally evaluate NAFNetSmall on a source-image-disjoint
SIDD partition and MobileNetV2~\cite{sandler2018mobilenetv2} on CIFAR-10. For restoration, 160 scene-instance
directories are split into 128 training, 16 validation and 16 test instances
using seed 9101, before four-patch expansion. No source image supplies
patches to multiple splits. This is \emph{scene-instance-disjoint}, not
physical-scene-disjoint: there are ten physical-scene IDs in training,
seven in validation and eight in testing. All eight test scene IDs also
occur in training, and six occur in validation. The experiment removes
source-image patch overlap but does not test unseen physical scenes.

Restoration models are trained for 50 epochs with AdamW at initial learning
rate $2\times10^{-4}$ and batch size eight. NAFNetSmall uses width 24,
encoder block counts $(2,2,4)$, two middle blocks and decoder block counts
$(2,2,2)$. MobileNetV2 uses width multiplier one, a 45,000/5,000/10,000
CIFAR-10 train/validation/test split (seed 9201), and 120 training epochs
with SGD at initial learning rate 0.05, momentum 0.9 and weight decay
$5\times10^{-4}$. The comparisons use six requested BOP targets, a
2,000-proposal refinement budget and one LSQ+ replicate with base seed 42 (actual seed $42+1000t$
for zero-based target index $t$). Restoration
training uses L1 loss and classification uses cross-entropy; global
activation profiling compares against FP32 outputs as described above.

\begin{table}[tbp]
\centering
\small
\setlength{\tabcolsep}{5pt}
\caption{Additional architectures at a requested BOP target of 4.102\%.
Actual routed-layer BOPs differ. Restoration cells show PSNR (dB) / SSIM;
classification cells show top-1 accuracy (\%). Results are single recovery
runs. SIDD splits are scene-instance-disjoint with shared physical scenes.}
\label{tab:generalization}
\begin{tabular}{lrrr}
\toprule
Method & Achieved BOPs (\%) & Static quality & LSQ+ quality \\
\midrule
\multicolumn{4}{l}{\textit{NAFNetSmall / SIDD (PSNR / SSIM)}} \\
FP32 reference & 100.000 & 34.743 / 0.8710 & -- \\
Uniform & 4.102 & 31.988 / 0.7895 & 32.865 / 0.8134 \\
HAWQ-style & 4.111 & 29.108 / 0.7483 & 33.571 / 0.8461 \\
QUBO & 4.113 & 29.407 / 0.7620 & 33.703 / 0.8429 \\
QUBO + PROTES & 4.107 & 33.593 / 0.8317 & 34.079 / 0.8514 \\
\midrule
\multicolumn{4}{l}{\textit{MobileNetV2 / CIFAR-10 (accuracy, \%)}} \\
FP32 reference & 100.000 & 91.720 & -- \\
Uniform & 4.102 & 91.220 & 91.770 \\
HAWQ-style & 3.580 & 90.450 & 91.760 \\
QUBO & 3.493 & 90.460 & 91.680 \\
QUBO + PROTES & 3.475 & 90.100 & 91.860 \\
\bottomrule
\end{tabular}
\end{table}

Table~\ref{tab:generalization} reports quality and achieved cost at one
requested target for each architecture. On NAFNetSmall, quality changes from 33.703 to 34.079 dB at approximately
4.11\% BOPs. On MobileNetV2, refinement reduces static accuracy from
90.46\% to 90.10\% but improves post-LSQ+ accuracy from 91.68\% to
91.86\%. These outcomes argue against a universal refinement benefit.

Across all six targets, the highest recorded scores are 34.269 dB for NAFNetSmall
(uniform static, 6.25\%), and 91.89\% for MobileNetV2 (uniform LSQ+,
3.418\%). These maxima have different costs and are not matched-budget
comparisons. Repeated requested targets can map to the same route; recovery
variation must not be interpreted as independent routing-seed uncertainty.

\subsection{Observed Quality--Compute and Quality--Storage Frontiers}
\label{sec:final_pareto_frontier}

We evaluate the routing strategies under two quantization settings: raw
Static Symmetric quantization and LSQ+ fine-tuning.

\subsubsection{Weight-Only Quantization: Quality--Storage Trade-offs}

We first isolate weight quantization and evaluate PSNR against the weight
storage compression ratio.

\begin{figure}[tbp]
\centering
{\small\textbf{(a) Weight-only allocation: quality versus storage}}\par
\smallskip
\includegraphics[width=\textwidth]{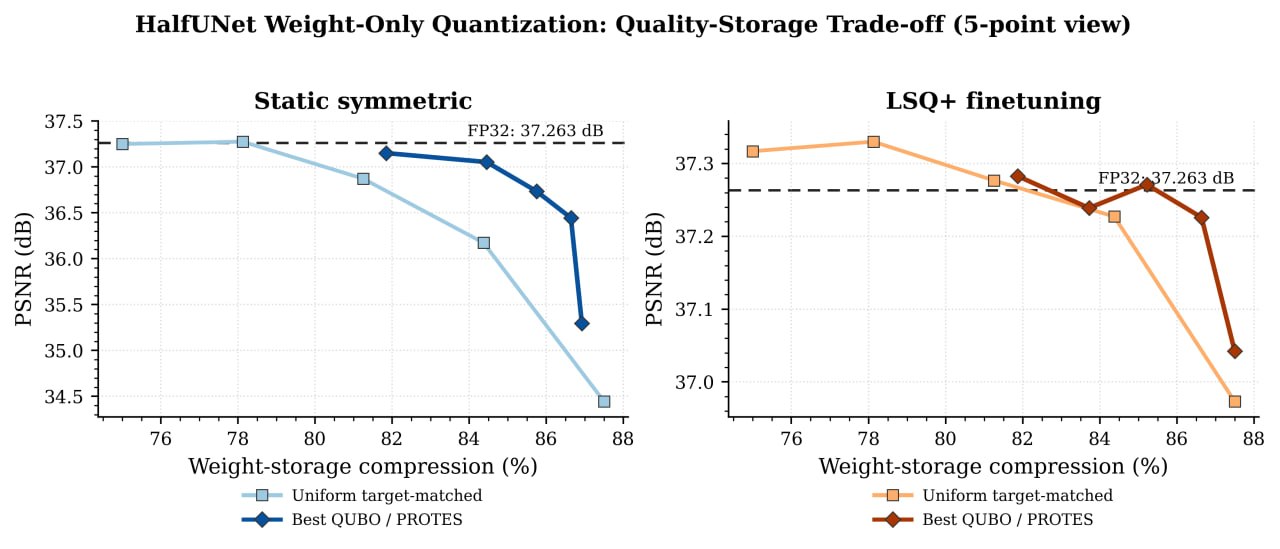}
\par\medskip
{\small\textbf{(b) Joint weight--activation allocation: quality versus compute}}\par
\smallskip
\includegraphics[width=\textwidth]{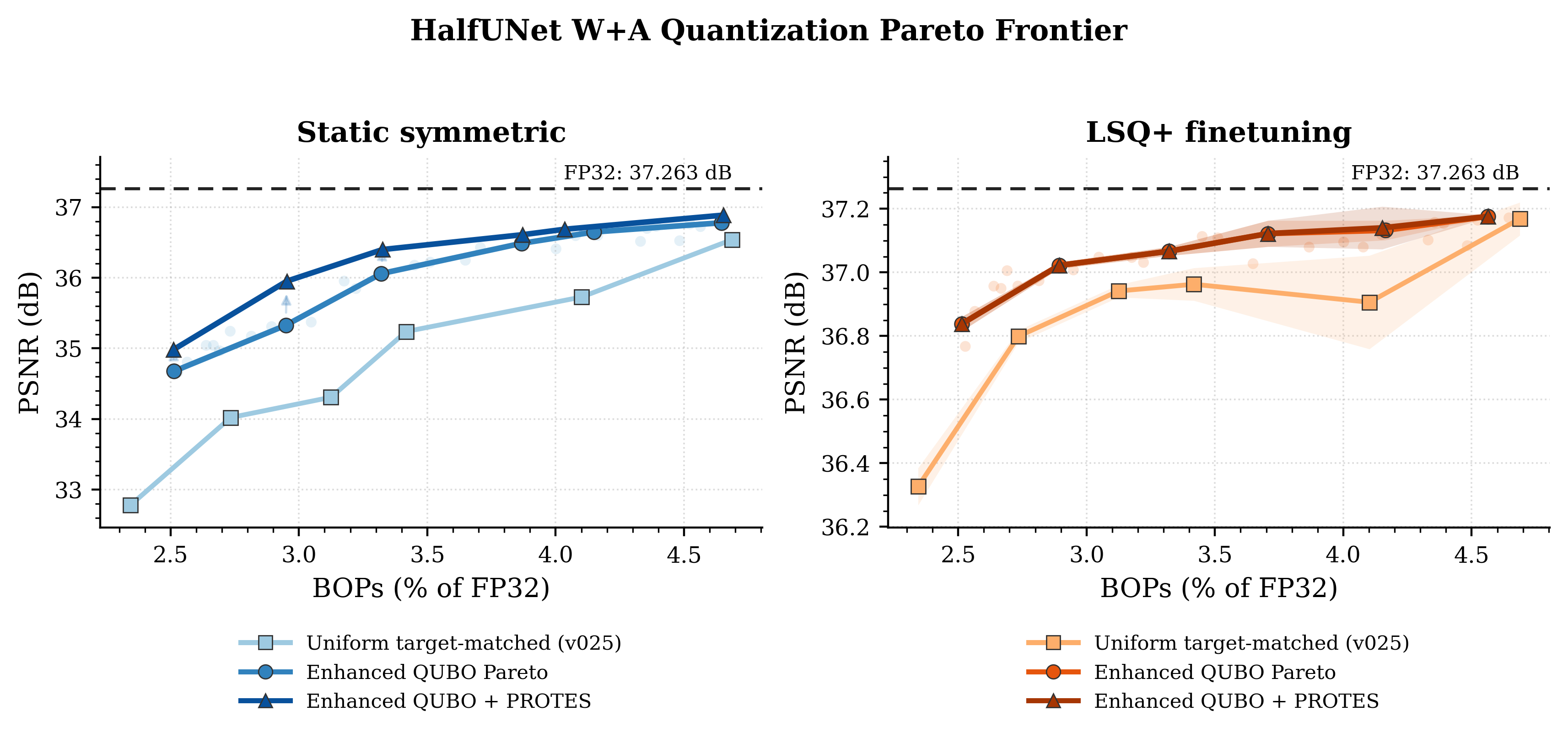}
\caption{Saved HalfUNet trade-off views, preserved as a combined
appendix figure. Each row shows static symmetric quantization (left) and
LSQ+ fine-tuning (right). Row (a) uses weight-storage compression; row (b)
uses the normalized BOP proxy, so their horizontal axes are distinct. The
weight-only legend reports the best observed QUBO/PROTES configurations;
the joint view separates QUBO and refined routes and retains the richer
original dense-sweep image. These selected-point views provide preliminary context and are separate from the six-target records
in main Figure~\ref{fig:wa_frontier}. The shading in row (b) is preserved from the supplied
image; its exact band definition and run counts per point are unverified
for this plot. It is not interpreted as a confidence interval. Connecting
lines and faint markers do not establish additional independent routing runs.}
\label{fig:weight_only_pareto}
\label{fig:original_frontiers}
\end{figure}

As illustrated in Fig.~\ref{fig:original_frontiers}(a), at extreme compression
rates approaching 87.5\%, the symmetric uniform baseline falls to
approximately 34.5 dB, while the best observed QUBO/PROTES configurations
remain above 35.0 dB at comparable compression.

With LSQ+ fine-tuning, the performance gap is reduced, but sensitivity-aware
routing remains competitive with the uniform baseline at the most aggressive
compression levels.

\subsubsection{Joint Weight and Activation (W+A): Quality--Compute
Trade-offs}

For joint weight--activation quantization, we evaluate PSNR against the
normalized BOPs compute proxy. Figure~\ref{fig:original_frontiers}(b) retains
the original dense-sweep W+A plot alongside the weight-only view. Keeping the
rows separate preserves their distinct horizontal axes and method legends.

Table~\ref{tab:main_comparison} reports the six requested operating points
with their achieved BOP ratios and both static and recovered quality.
The original appendix plot uses a different selected-point view; its curves
are not additional independent evaluations of those six-target records.
Main Figure~\ref{fig:wa_frontier} remains the visual comparison tied directly
to that table.

\paragraph{Development joint-quantization observations.}
The preliminary dense sweep records an approximate uniform static score of 32.8 dB
near 2.4\% BOPs, approximately 34.7 dB for enhanced QUBO, and approximately
35.0 dB for refinement near 2.5\%. After LSQ+ it records approximately
36.8 dB for the refined route versus 36.3 dB for uniform quantization near
that boundary. We retain these preliminary observations separately from the
six-target comparison: they have different operating points and should
not be pooled into the latter's rows or treated as additional independent
runs. These approximate observations are not confidence intervals.

\subsection{Ablation Studies: Negative Results}
\label{sec:negative_results}

We additionally evaluated several alternative sensitivity and routing
strategies:

\begin{itemize}
    \item \textbf{Second-Order Hessian Trace Estimation:} At $\gamma=200.0$,
    a Hessian-aware QUBO using Hutchinson's estimator underperformed the
    first-order sensitivity formulation in the evaluated setting.

    \item \textbf{LSQ+ Micro-Calibration:} Optimizing the step size using 50
    Adam iterations on isolated tensors before MSE evaluation resulted in a
    PSNR of 33.427 dB in that preliminary configuration. The expanded
    experiments nevertheless use a calibrated-weight-error variant; this
    single result does not establish that calibration is generally inferior.

    \item \textbf{Sub-Layer Block-Wise Routing:} Channel-group routing based
    on sequential slicing, logical variance sorting, and activation-correlation
    clustering performed poorly in the evaluated experiments and increased
    optimization complexity.

    \item \textbf{Dynamic Distributional Entropy:} Using activation entropy
    to unlock aggressive $\{4,5\}$-bit choices caused substantial degradation
    in the tested configurations, indicating that activation sparsity alone was
    not a reliable indicator of quantization robustness.

    \item \textbf{Anti-Correlation Splitting:} Round-robin distribution of
    correlated channels across logical blocks produced unstable results in the
    evaluated experiments.
\end{itemize}

These results should be interpreted as empirical comparisons among the specific
alternatives evaluated in this study rather than as evidence that the tested
alternatives are universally inferior.

% ============================================================================
% 7. CONCLUSION
% ============================================================================

\end{document}